%% file: main.tex
\documentclass[lettersize,journal]{IEEEtran}

\usepackage{amsmath,amsfonts}
\usepackage{algorithmic}
\usepackage{algorithm}
\usepackage{array}
\usepackage[caption=false,font=normalsize,labelfont=sf,textfont=sf]{subfig}
\usepackage{textcomp}
\usepackage{stfloats}
\usepackage{url}
\usepackage{verbatim}
\usepackage{graphicx}
\usepackage{cite}
\usepackage{booktabs} 
\usepackage{xcolor}
\newcommand{\faSnowflake}{\raisebox{-0.125em}{\includegraphics[height=1em]{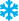}}}
\newcommand{\faFire}{\raisebox{-0.125em}{\includegraphics[height=1em]{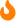}}}
\usepackage{multirow}
\usepackage{gensymb}
\usepackage{standalone}
\usepackage{booktabs}
\usepackage{tikz}
\usepackage{makecell}
\usepackage[table]{xcolor} % pour colorer les entrées du tableau
\usepackage[
    colorlinks=true,
    linkcolor=blue,
    citecolor=blue,
    urlcolor=blue
]{hyperref}
\newcommand{\orcidlink}[1]{\texorpdfstring{\href{https://orcid.org/#1}{\includegraphics[height=\fontcharht\font`X]{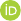}}}{}}
\usepackage{cleveref}

\usepackage{xcolor} %remove, used for macros
\usepackage{xspace}

\definecolor{bestgreen}{HTML}{ace2ae}   
\definecolor{secondgreen}{HTML}{E8F5E9} 
\definecolor{redlight}{HTML}{fcbdbd} % rouge clair 

\crefname{figure}{Figure}{Figures}
\crefname{table}{Table}{Tables}
\crefname{section}{Section}{Sections}
\crefname{equation}{Equation}{Equations}

\newcommand{\cmark}{\textcolor{green!60!black}{$\checkmark$}}
\newcommand{\xmark}{\textcolor{red!70!black}{$\times$}}
\newcommand{\greymidrule}{%
  \arrayrulecolor{gray!50}\midrule\arrayrulecolor{black}%
}
\newcommand{\grouprule}[1]{\addlinespace[1pt]\cmidrule(lr){#1}\addlinespace[1pt]}
\newcommand{\avgv}[1]{\textcolor{black!55}{#1}}

\newcommand{\ra}{$\rightarrow$}  
\newcommand{\lra}{$\leftrightarrow$}

\newcommand{\modelname}{$\mathrm{SEMIS}$\xspace}  
\newcommand{\modelnamemath}{\mathrm{SEMIS}\xspace}  
\newcommand{\modelnameFS}{$\mathrm{SEMIS}$-$\mathrm{FS}$\xspace}

\newcommand{\modelnamein}{$\mathrm{SEMIS}^\mathrm{I}$\xspace}

\newcommand{\inera}{$\mathrm{SEMIS}^{\mathrm{I}}_{\mathrm{era5}}$\xspace}

\newcommand{\modelnameout}{$\mathrm{SEMIS}^\mathrm{T}$\xspace}

\newcommand{\outera}{$\mathrm{SEMIS}^{\mathrm{T}}_{\mathrm{era5}}$\xspace}
\newcommand{\outtopo}{$\mathrm{SEMIS}^{\mathrm{T}}_{\mathrm{topo}}$\xspace}
\newcommand{\outsoil}{$\mathrm{SEMIS}^{\mathrm{T}}_{\mathrm{soil}}$\xspace}
\newcommand{\outcombo}{$\mathrm{SEMIS}^{\mathrm{T}}_{\mathrm{combo}}$\xspace}
\newcommand{\specdataname}{SPAN\xspace} % car la canicule me rend zinzin
\newcommand{\cropdataname}{RapidCrops$^{SG}$\xspace} 

\newcommand{\dvan}{$\Delta_{\modelnamemath}$\xspace}

\newcommand{\imp}[1]{\emoji{grinning-squinting-face}}
\newcommand{\ignore}[1]{}

\begin{document}
%\showthe\columnwidth

\title{Spatial Generalization of Satellite Image Time Series Foundation Models: Role of Auxiliary Variables as Inputs or Targets in Vegetation Type Mapping}

\author{Sarah~Brood\,\orcidlink{0009-0004-2609-6933},
        Iris~Dumeur\,\orcidlink{0009-0000-0182-2004},
        J\'er\'emy~Anger\,\orcidlink{0009-0007-3319-6037},
        Aur\'elien~de~Truchis\,\orcidlink{0000-0003-3515-4176},
        Philippe~Ciais\,\orcidlink{0000-0001-8560-4943},
        and~Alexandre~d'Aspremont\,\orcidlink{0000-0003-3851-216X}% <-this % stops a space

%\thanks{Manuscript received XX; revised XX.}% <-this % stops a space
\thanks{This work was supported in part by the French National Research Agency (ANR) through the AI4Forest project (ANR-22-FAI1-0002).}
\thanks{S. Brood is with École Normale Supérieure, INRIA, CNRS, PSL Research University, 75005 Paris, France (e-mail: sarah.brood@ens.fr). \textit{(Corresponding author: Sarah Brood.)}}% <-this % stops a space
\thanks{I. Dumeur and J. Anger are with Universit\'e Paris-Saclay, CNRS, ENS Paris-Saclay, Centre Borelli, 91190 Gif-sur-Yvette, France.}% <-this % stops a space
\thanks{A. de Truchis is with Kayrros SAS, 75009 Paris, France.}% <-this % stops a space
\thanks{P. Ciais is with the Laboratoire des Sciences du Climat et de l'Environnement, LSCE/IPSL, CEA-CNRS-UVSQ, Universit\'e Paris-Saclay, 91191 Gif-sur-Yvette, France.}% <-this % stops a space
\thanks{A. d'Aspremont is with CNRS - École Normale Supérieure, PSL Research University, INRIA, 75005 Paris, France, and also with Kayrros SAS, 75009 Paris, France.}}

\markboth{IEEE Transactions on Geoscience and Remote Sensing}%
{Brood \MakeLowercase{\textit{et al.}}: Spatial Generalization of SITS Foundation Models}% <-this % stops a space


\maketitle

%###################%

\begin{abstract}
\input{sections/0_abstract}

\end{abstract}


%###################%

\section{Introduction}
\input{sections/1_intro}

\section{Related Work}
\input{sections/2_related}

\section{Method}
\input{sections/3_method}

\section{Data}
\input{sections/4_data}

\section{Experimental Set-up}
\input{sections/5_xp}

\section{Results}
\input{sections/6_results}

\section{Conclusion}
\input{sections/7_conclusion}

\section*{Acknowledgment}% -> hide durring submission for anonymity
This work was granted access to the HPC resources of IDRIS under the allocations 2026-AD010117360 and 2026-AD010114718R2 made by GENCI. 

The authors thank Ewan Sean and Hugo Rechatin for downloading the Sentinel-2 L2A data, and Martin Schwartz for his support throughout this work.

This article contains modified Copernicus Sentinel data (2017 to 2024). 
The climatic variables were generated using Copernicus Climate Change Service information.

The authors used Claude (Anthropic) to assist with writing and debugging parts of the code for the model and the data analysis, and with drafting the drawing code for the schematic diagrams in \cref{fig:backbone,fig:aux_in_int,fig:aux_targ_int,fig:pretraining,fig:protocol,fig:local_att}, which were subsequently revised by the authors. Claude was also used to improve the English grammar and style of the manuscript. All AI-assisted code was reviewed and tested, and all results were obtained and verified by the authors.

%\section*{References}
\bibliographystyle{IEEEtran}
\bibliography{refs/data,refs/deep,refs/mapping,refs/climate_refs}

%###################%
\appendix
\input{sections/X_appendix}

%###################%

\end{document}

%% file: sections/0_abstract.tex
Large-scale tree species and crop type mapping is strongly limited by the scarcity of labeled data.
Foundation models (FMs) for Satellite Image Time Series (SITS) offer new opportunities by providing easy-to-use representations.
However, whether these representations generalize geographically well enough for large-scale vegetation map production has not been sufficiently evaluated.
In addition, a growing number of models integrate auxiliary variables, which could partly inform them about intra-class spatial variability, yet their use as inputs and as pre-training targets has rarely been compared.
In this study, we introduce a benchmark that assesses spatial generalization over Europe on crop type and tree species mapping, and covers 18 transfers between spatial domains.
On this benchmark, we first compare four existing SITS FMs in a realistic spatial generalization setting, where the pre-trained encoders are kept frozen. 
Second, we propose SEMIS, a reference backbone used to evaluate the integration of auxiliary variables as inputs or as pre-training targets.
Our results show that all models exhibit a large performance drop when tested on an unseen region and that, for crop types, this drop increases with the distance between regions.
They also suggest that auxiliary variables used as inputs tend to give the best in-domain scores, whereas, when used as pre-training targets, they tend to give the best cross-domain scores.
However, our in-depth analysis shows that these results vary with the auxiliary variables used and with the pair of source and target regions considered.

% current word ccount: 245
% https://www.latexwordcounter.com/

%% file: sections/1_intro.tex
\label{sec:intro}

%% enjeu cartes vegetation
% construire carte vegetation crucial pour climate change mitigation et autres
% tree species maps (biomass, mangement etc)
% crop types (food etc)
\IEEEPARstart{L}{arge-scale} vegetation type mapping supports climate change mitigation and adaptation, biodiversity conservation, and food security. For instance, tree species maps at regional to continental scale are a key input for biomass and carbon accounting, climate-adapted forest management, and biodiversity monitoring~\cite{forest_sink,ipcc_2023,gamfeldt2013higher}. 
Similarly, crop type maps support agricultural land-use monitoring, yield forecasting, and food security assessment~\cite{WEISS2020111402, porter2014food}.

%% S2 SITS
% données S2: open, couverture globale, multi spectral pour vegetation forte revisite %> SITS
% SITS %> phenology patterns 
% iregular, spectro spatial temp
%% S2 SITS
% données S2: open, couverture globale, multi spectral pour vegetation forte revisite %> SITS
% SITS %> phenology patterns 
% iregular, spectro spatial temp
Producing such large-scale vegetation maps relies on frequent satellite acquisitions, as temporal patterns are essential to distinguish between vegetation types.
Among the available sensors, Sentinel-2~\cite{sentinel2} provides open data combining global coverage, multi-spectral acquisitions including infrared bands sensitive to vegetation, and a revisit time of about five days. The resulting data, referred to as Satellite Image Time Series (SITS), are complex four-dimensional objects with spectral, spatial, and temporal dimensions. In addition, SITS are by nature irregular and may be affected by clouds.
%On the other hand, many tree species and crop types are better distinguished by these temporal patterns than by a single acquisition~\cite{grabska2019forest}. This is why SITS are of high interest in order to capture vegetation temporal patterns.
%However, SITS acquisition are irregular, cloud cover making the number and dates of valid acquisitions vary across regions.
%This irregularity, combined with their spectral, spatial, and temporal dimensions, makes SITS complex to process.

%% pourquoi transferer 
% carte mainly a echelle nationale 
% crops: dataset %> cross continent 
% forests: 
	% encore moins de labels forest inventories, pas revisit annuelles, protocoles et class différentes 
% donc cartes continentales ou nationales si pas de labels: demande de re%utiliser des labels entre regions %> appuyer sur manque de labels
In practice, due to a lack of labels over large areas, tree species and crop type maps are mainly produced up to national scale~\cite{grabska2024map,breidenbach2021national,welle2022mapping,blickensdorfer2022mapping,blickensdorfer2024national}. 
This label constraint is particularly critical for tree species mapping, for which no harmonized labeling system exists. 
More precisely, tree species annotations mainly come from national forest inventories, which rely on costly field campaigns, are revisited only every few years, and follow national protocols~\cite{grabska2024map,gaydon2025pureforest,ahlswede2022treesatai}. 
For crops, some harmonized label datasets exist at the continental scale, for example in Europe through the Land Parcel Identification System (LPIS)~\cite{schneider2023eurocrops,rapidcrops}. 
Emerging initiatives try to provide labels across continents~\cite{van2023worldcereal}, although large parts of the world are still not covered.

Furthermore, while vegetation labels are scarce, unlabeled Sentinel-2 SITS are abundant.
To take advantage of these data, an increasing number of Earth observation (EO) foundation models (FMs) have been proposed, including several FMs designed for SITS~\cite{tseng2023lightweight,dumeur2025paving,feng2025tessera,astruc2025anysat,brown2025alphaearth,herzog2026olmoearth}.
These models rely on an encoder pre-trained once with a self-supervised learning strategy to produce informative and generic representations of the input SITS. In particular, this field of research is now moving toward the production of easy-to-use SITS representations~\cite{dumeur2025paving,feng2025tessera,brown2025alphaearth}, meaning that end users (e.g., geoscientists or agricultural agencies) should be able to use these representations as is. 
Consequently, once pre-trained, the model should be used frozen, and only a shallow classifier should be trained on the output SITS representations using the available labels. 
In line with this, recent FMs are even starting to distribute ready-to-use SITS embeddings~\cite{feng2025tessera,brown2025alphaearth}. 
Therefore, for large-scale vegetation mapping under strong label scarcity, the output SITS representations should be spatially generic. 
In other words, a shallow classifier trained on SITS representations from one area should ideally be applicable to another area, where labels are not available for training. 
We refer to this property as the spatial generalization of representations.
In contrast, unsupervised domain adaptation techniques adapt the model to a single target domain, which must be known in advance.
In this study, we consider that FMs should provide representations that can be applied to numerous and varied areas.

However, whether SITS FM representations are spatially transferable has rarely been evaluated.
Existing evidence on the spatial generalization drop comes from two lines of work. %which only partly address this measurement.
On the one hand, existing evaluations on the spatial generalization of EO FMs suggest that these models suffer a performance drop on unseen regions~\cite{doerksen2026earthshift,marsocci2024pangaea,gordon2026mmearth}.
Yet, these evaluations rely on single-date imagery, and most of them train part of the model on source labels by fine-tuning the encoder~\cite{gordon2026mmearth} or require training a deep decoder on top of the frozen model~\cite{marsocci2024pangaea}, which does not allow us to isolate the spatial generalization of the representations.

On the other hand, spatial shift on SITS has mostly been studied with fully supervised models trained from scratch~\cite{nyborg2022timematch,orynbaikyzy2022spatial,sykas2022sentinel}.
These studies also report a drop across regions that is consistent with the intra-class SITS variability reported for vegetation across regions, caused by differences in climate, soil, topography, and management, which affect phenology~\cite{nyborg2022timematch,koch2025assessing,grabska2021evaluating,bajocco2021exploring}.

Across both lines of work, most evaluations report scores aggregated over transfers or over classes~\cite{nyborg2022timematch,doerksen2026earthshift,marsocci2024pangaea,gordon2026mmearth}; such scores do not show the patterns of the drop, that is, how it varies at both domain and class levels. 
For instance, they do not indicate whether some source-target transfers are harder than others, or whether a class-level drop is shared across all transfers.

It therefore remains unclear where this drop occurs and which drop patterns it follows.
We posit that this lack of spatial generalization evaluation is partly due to the absence of harmonized labeled datasets, which, to our knowledge, do not exist for tree species and are only very recently available for crops.
%Analyzing it also requires datasets with labels harmonized across regions, which exist for crops but, to our knowledge, not for tree species.

%% auxiliaries 
% labels cibles manque mais pas auxiliaires environmentales maps partout !
% FMs les integre en input, target, les deux, citer mais pas qui quoi comment
% en entrees requise a l'inference, en cible non
% in%n%out results
% gap: FMs ne compare pas ces choix, pas d'ablation/comparaisons, surtout en regime frozen, backbone constant et sans données cibles. 
Beyond measuring the spatial generalization drop, one may also try to mitigate it by providing the model with information about its causes.
Among the causes of intra-class SITS variability across regions, climate, soil properties, and topography can be described by environmental variables.
Unlike target labels, these variables, denoted in this paper as auxiliary variables, are available over any region as geo-referenced layers, independently of the SITS.

In addition, several FMs already integrate auxiliary variables, either as additional pre-training targets only~\cite{nedungadi2024mmearth,brown2025alphaearth}, or as both inputs and targets~\cite{tseng2023lightweight,tseng2025galileo}.
When used as inputs, they are encoded alongside the spectral information and must be available at inference.
In contrast, when used as targets, they are predicted as an additional pre-training objective and are not required at inference.
Xie et al.~\cite{xie2021innout} compared these two integrations with the In-N-Out study.
In this study, auxiliary inputs improved in-distribution accuracy but could reduce out-of-distribution accuracy, whereas auxiliary targets improved out-of-distribution accuracy.
However, this analysis relied on task-specific models: auxiliary inputs were fed to a supervised model, and auxiliary targets were used to pre-train on unlabeled data that included out-of-distribution samples.
To our knowledge, the two integrations have not been compared within a frozen SITS FM without target-domain data.
It therefore remains unclear how integrating auxiliary variables as inputs or as targets affects the spatial generalization of SITS FM representations.

%% Nous 
% Questions: 
	%Therefore, this study addresses two questions: 1) Where does this performance drop occur under spatial shift, and what does it depend on? 
    %2) Does auxiliary integration reduce this gap, and does it matter whether the auxiliaries are integrated as inputs or as targets?
% Explication protocole: pas de data target, 25 paires ! 
% Benchmark  datasets on 2 tasks (mais pas trop précis)
% evaluate baseline models
% evaluate backbone semis + variants 
Therefore, this study addresses two questions: (1) Among existing pre-trained SITS encoders~\cite{tseng2023lightweight,dumeur2025paving,astruc2025anysat,feng2025tessera}, where does the performance drop occur when the training and test spatial domains vary, and which patterns does it follow? (2) Does auxiliary variable integration mitigate these patterns, and does it matter whether auxiliary variables are integrated as inputs or as targets?
To address these questions, we evaluate the spatial generalization of frozen SITS encoders on two tasks: tree species and crop type mapping.
The evaluation consists of a linear classifier trained on frozen SITS representations from a source domain and tested on different target domains, without any target data for training or model selection, over 18 source/target region pairs.
We evaluate four existing SITS FMs and propose a control encoder, named \modelname, designed to compare the two auxiliary variable integrations within the same backbone, through two variants that use auxiliary variables either as inputs (\modelnamein) or as targets (\modelnameout). 
%% contribs 
% SPAN et RapidCropsS
	% 2 taches, labels unifiés mesure shift 
		%  first arbre: spatial, avec interannuel %> cas d'usage
		% crops: purement spatial,
% protocole spatial generalisation evaluation de 4 FMs, avec analyse par pair, class et 
% SEMIS pour comparer avec integration aux
	% encodeur spectro%spatio%temporel, représentation de taille fixe, pré%entraîné en MAE temporel, control backbone 
The contributions of this paper can be summarized as follows:
\begin{itemize}
    \item We introduce two multi-domain datasets with harmonized labels to measure the spatial generalization:
    \begin{itemize}
        \item \specdataname, a tree species mapping dataset aggregating three national datasets~\cite{grabska2024map,ahlswede2022treesatai,gaydon2025pureforest} with stand-level labels, across Europe under six shared classes. %Its domains also differ in acquisition year, as in operational usage.
        \item \cropdataname, a crop type mapping dataset derived from RapidCrops labels~\cite{rapidcrops} on four Sentinel-2 tiles in Denmark, Austria, and France (two tiles) as in TimeMatch~\cite{nyborg2022timematch}. % where the shift is purely spatial.
    \end{itemize}
    \item We conduct an in-depth evaluation of the spatial generalization capacity of four existing SITS FMs~\cite{tseng2023lightweight,dumeur2025paving,astruc2025anysat,feng2025tessera} under a realistic spatial generalization protocol.
    \item We design \modelname (SITS prEtrained Mae tImeSformer), a spatio-temporal masked-autoencoder encoder that maps Sentinel-2 SITS to fixed-size representations and serves as a reference backbone to compare the two auxiliary variable integrations under similar settings.
    %\item We design \modelname, a spatio-temporal encoder that maps Sentinel-2 SITS to fixed-size embeddings and is pre-trained as a masked autoencoder over acquisitions. It serves as a control backbone to compare the two auxiliary integrations with the same architecture, pre-training data, and schedule, in the frozen regime.
     %\item We propose a single-source spatial generalization protocol for frozen encoders, which attributes each performance drop to a source/target pair. We apply it to four SITS FMs and break the results down by domain pair and by class, showing that aggregated cross-domain scores hide differences at both levels.
\end{itemize}
%% hors bullet
%  le resultat du benchmark pour le drop? 
% le results de input vs target ?
Our results first demonstrate that the spatial generalization performance drop varies across domain pairs and classes. For crop types, it grows with the distance between domains, whereas for tree species, the drop remains large even between the two closest regions. Second, auxiliary variables used as pre-training targets tend to give the best spatial generalization performance without being required at inference, whereas used as inputs they mainly improve in-domain scores.
These gains, however, often vary with the regions used as training or test sets. 

%% organisation du papier 
This article is organized as follows: \cref{sec:related} reviews the related work, \cref{sec:method} presents \modelname, its pre-training, and its auxiliary integration, \cref{sec:data} describes the datasets and input data, \cref{sec:xp} details the spatial generalization protocol and the baseline encoders, \cref{sec:results} reports the results, and \cref{sec:conclu} concludes.

%% file: sections/2_related.tex
\label{sec:related}

% intro de la section: 3 sub, dire qu'on se concentre sur S2 SITS et citer pas SITS quand travail avec SITS inexistent ?
% > il faut que ce soit clair pour le lecteur que ce pb de généralisation géographique est très peu considéré pour les modèles exploitant des SITS

This section is divided into three parts. We first review EO FMs that use SITS and how they integrate auxiliary variables (\cref{sec:rel_fm_aux}). We then examine how spatial generalization is currently evaluated (\cref{sec:rel_eval}), and finally, we discuss the strategies proposed to improve spatial generalization (\cref{sec:rel_addressing}).
 % We focus on studies based on SITS, and refer to single-date studies only when few or no SITS studies are available.

\subsection{Earth observation foundation models for SITS} 
\input{sections/tables/related_encoders}
\label{sec:rel_fm_aux}
% A. SITS FMs: sits embedding exists
% > insister sur ceux qui proposent des représentations de taille fixe
% ways to use them 2. archis: cnn att, full att, (regarder qui fait full att) en parallel en video on utulise full att timesformer
% 1. temporal vs spatial from garnot~\cite{garnot2019time}
% 2. temporal only, pixel-wise: tessera, presto, sits-bert
%  3. spatial context: anysat, alphaearth, alise
% > ton papier ne considère que le mono-modal, je ne vois pas trop l'intérêt d'évoquer autant le multimodal
%The first approach uses convolutions to integrate spatial context, then processes the temporal dimension with attention~\cite{yuan2022sits,dumeur2025paving}. 
%A second group integrates both spatial and temporal context. These models follow two main approaches. 
%The second approach uses spatio-temporal attention, mainly found in works that focus on multi-modal integration~\cite{astruc2025anysat,tseng2025galileo,brown2025alphaearth}. 
%  efficient 
%  critiquer heavy space time archis  finir pas TImeSformer 
%  4. ready to use: tessera, alphaearth: ppl don't train models + different ways to give fixed size: TESSERA: pooling vs ALISE pour fixed size temporal
%  5. SSL strategies (mae etc)
%
%Masked reconstruction, following MAE~\cite{he2022masked}, is also a common objective. 
%Masking entire acquisitions appears well suited to SITS: it simulates missing dates or cloud cover, and it may encourage the model to learn temporal variations.
Several Earth observation FMs have recently been proposed to produce representations from SITS that can be used for many downstream tasks.
These models differ in their self-supervised learning (SSL) objectives, their architectures, and their use of auxiliary data.
\Cref{tab:related_fm} summarizes these characteristics for representative encoders, including the four baselines evaluated in this study and our models.
First, in SSL, reconstruction is a widely adopted pretext objective~\cite{tseng2023lightweight,dumeur2025paving,brown2025alphaearth,cong2022satmae,szwarcman2025prithvi,Labatie_2026_WACV,irma_2026}. Other techniques, such as joint-embedding~\cite{astruc2025anysat,feng2025tessera,tseng2025galileo} or discriminative objectives~\cite{dumeur2025paving,brown2025alphaearth,tseng2025galileo,astruc2025anysat}, may also be employed, and these are often combined with a reconstruction objective within the same model.
%First, in SSL, while numerous methods employ reconstruction~\cite{tseng2023lightweight,dumeur2025paving,brown2025alphaearth}, other techniques may also be employed such as joint-embedding~\cite{astruc2025anysat,feng2025tessera,tseng2025galileo}, or discriminative objectives~\cite{dumeur2025paving,brown2025alphaearth,tseng2025galileo,astruc2025anysat} and these are often combined within the same model.

Second, these models also differ in how they process the temporal and spatial dimensions of the SITS. For vegetation mapping tasks such as crop type mapping, temporal features have been found more informative than spatial ones, although some spatial context remains beneficial~\cite{garnot2019time}. Accordingly, a first group of encoders processes the time series of each pixel independently, without spatial context~\cite{feng2025tessera,tseng2023lightweight,yuan2020self}.
Other models integrate spatial context, either by factorizing the spatial and temporal dimensions, processed sequentially~\cite{dumeur2025paving,astruc2025anysat} or in parallel~\cite{brown2025alphaearth}, or by using joint attention over all spatio-temporal tokens~\cite{tseng2025galileo,herzog2026olmoearth,szwarcman2025prithvi}.
In theory, spatio-temporal attention architectures from the natural video field offer high flexibility and capacity, but they are limited by their computational cost. 
Spatio-temporal attention scales quadratically with the number of tokens, and therefore with both the number of dates and the spatial extent. To reduce this cost, TimeSformer~\cite{bertasius2021space} divides space-time attention for natural videos, applying temporal attention and then spatial attention within each encoder block.
In addition, several studies advocate that SITS-based FMs should produce representations that are ready to use by end users~\cite{brown2025alphaearth,dumeur2025paving,feng2025tessera}. 
Consequently, FMs should be kept frozen in downstream tasks and output fixed-size representations from irregular input SITS. Existing methods meet this latter requirement with different techniques. Some models apply temporal pooling at the downstream stage~\cite{tseng2023lightweight, tseng2025galileo,herzog2026olmoearth}, while others apply it during pre-training~\cite{feng2025tessera,brown2025alphaearth,astruc2025anysat} so that the temporal dimension is collapsed to one. Another work relies on cross-attention with learnable queries to produce fixed-size representations~\cite{dumeur2025paving}. This study also found that, in a reconstruction self-supervised setting, reducing the temporal dimension to one severely degrades SITS representations.

Third, some FMs also integrate auxiliary variables, while others rely on remote sensing observations only.
%We exclude from this definition acquisition dates, which are inherent to SITS.
%Geographic coordinates, even if they are encoded by several FMs~\cite{szwarcman2025prithvi,wang2025towards,tseng2023lightweight}, are also excluded because they have been reported to limit generalization to unseen regions~\cite{meyer2019importance}.
These auxiliary variables are used in three ways: only as inputs, as both inputs and targets, or only as targets. 
When used only as inputs, auxiliary variables provide additional information to the model.
To our knowledge, no SITS FM uses them only as inputs during pre-training.
However, this strategy is employed in several fully supervised SITS models~\cite{blickensdorfer2022mapping,blickensdorfer2024national,nyborg2022generalized}. 
%Auxiliary variables have also been used as inputs to downstream classifiers trained on frozen SITS FM embeddings~\cite{ball2026geospatial}.
In contrast, some SITS FMs use auxiliary variables both as inputs and as targets.
Presto~\cite{tseng2023lightweight} and Galileo~\cite{tseng2025galileo} encode them alongside the SITS, then randomly mask and predict them during pre-training.
Both models accept missing inputs at inference; however, to our knowledge, the effect of omitting auxiliary variables at inference has not been quantified. 
Conversely, auxiliary variables can be used as pre-training targets only, as in the SITS FM AlphaEarth~\cite{brown2025alphaearth} and in single-date models such as MMEarth~\cite{nedungadi2024mmearth}.
One advantage of this approach is that auxiliary variables are not required for downstream tasks.

% GAP
In summary, while auxiliary variables have been integrated into SITS FMs as inputs and targets, or as targets only, these strategies have not been fairly compared in a controlled setting, with the same architecture, SSL pre-training strategy, and downstream protocol with frozen parameters. %~\cite{brown2025alphaearth,feng2025tessera}. %Moreover, the effect of each integration strategy on spatial generalization has not yet been measured. 
Given all the above, we introduce \modelname, a mono-modal Sentinel-2 FM exploiting divided spatio-temporal attention~\cite{bertasius2021space} and producing fixed-size temporal representations~\cite{dumeur2025paving}.
Additionally, we develop two variants using auxiliary variables during pre-training, either as inputs (\modelnamein) or as targets (\modelnameout).

\subsection{Evaluating spatial generalization of EO FMs} 
\label{sec:rel_eval}
\input{sections/tables/related_gen}
% !!! existing shift exists but not measures on FMs SITS and anlsyis not developped
% A. CAUSES > parler rapidement des potentielles causes de ces shifts
Spatial generalization drop is documented for vegetation mapping, yet it has rarely been measured for SITS FMs. We first review its causes, then how it has been evaluated, and the datasets available to perform these measurements. 

%1. For trees, the reflectance of a species varies with site conditions, such as elevation and aspect, and with stand properties, such as age and crown closure. Elevation also delays the spring phenology of beech. This intra-species variability has also been observed with Sentinel-2 time series across Switzerland. 
% 2. crops: climate, soil -> phenology, crop calendar: timematch
%For crops, regional climate and soil conditions have been found to shape phenology~\cite{}, and climate shifts their calendar between regions. 
Remotely sensed signatures differ not only between classes, but also within a class, depending on their location~\cite{koch2025assessing}.
Site conditions, such as elevation, aspect, climate, and soil, affect both reflectance and phenology~\cite{grabska2021evaluating,potovcnik6568982interannual,bajocco2021exploring}, and climate shifts affect crop calendars and phenology between regions~\cite{nyborg2022timematch}. For trees, stand properties, such as age and crown closure, also add intra-class variability~\cite{grabska2021evaluating}.

% B. SHIFT MEASURED IN SUPERVISED MODELS  -> verifier qu'on est pas UDA
As a result, models trained in one region often lose accuracy when applied to another.
This drop has been measured for two kinds of models: fully supervised SITS classifiers and FMs using single dates or SITS.
\Cref{tab:related_spatgen} summarizes these studies according to their input data, model type, evaluation protocol, granularity of the reported results, and task.

% > il faut qu'on comprenne l'avancée par rapport à TimeMatch
% variabilité inter-annuelle (ball2026geospatial, verhulst2024temporal(trees))
%  overall accuracy dropped by 2.3 to 14.9 points across years for tree species~\cite{verhulst2024temporal} 
% and median F1 by 4 to 8 points across German regions for crop types~\cite{orynbaikyzy2022spatial}.
% FM embeddings have been evaluated for crop types within a region~\cite{lisaius2026towards}
% timematch
% donc il y a un shift mais on le aggregated score pas et c'est pas des FMs
% 1. crops: timematch avec SITS mais en UDA pas linear probing et pas des FMs
% level accuracy decreases when random forests are applied to other German regions~\cite{orynbaikyzy2022spatial}. 
% This drop remains when classes are harmonized across European region~\cite{nyborg2022timematch,sykas2022sentinel}.
For fully supervised SITS classifiers, accuracy drops have been measured for crop types, between regions of the same country and across European countries~\cite{orynbaikyzy2022spatial,nyborg2022timematch,sykas2022sentinel}.
% 2. forest: held out countries: miettinen2025pan
%  timber volume estimates degrade in countries excluded from training~\cite{miettinen2025pan}. For
% n MODIS time series when Africa is held out~\cite{xie2021innout}.
For land cover with MODIS time series, accuracy also decreased on a held-out region, Africa, for a model trained on the rest of the world~\cite{xie2021innout}.
% !  donc il y a un shift mais on le aggregated score pas et c'est pas des FMs
Yet, the spatial generalization evaluation of these fully supervised models is often limited. They rarely report the drop both per transfer direction and per class (\cref{tab:related_spatgen}). Such evaluations therefore cannot fully characterize spatial generalization drop patterns.
% are evaluated on few geographical regions, often in one transfer direction only. %and with one aggregated score.

% C. SHIFTS MEASURE in EO FMs : VERIFIER LINEAR PROBIGN OU PSA 
% 1.  mesurent le spatial shift : SINGLE DATE
%MMEarth-Bench uses Africa as a geographic held-out test region and concludes that geographic generalization remains poor~\cite{gordon2026mmearth}.
%EarthShift reports that FMs perform worse in out-of-distribution settings on average (including geographical shift)~\cite{doerksen2026earthshift}.
%PANGAEA includes one cross-region experiment, which also shows a strong drop\cite{marsocci2024pangaea}.
For single-date FMs, benchmarks have reported that their accuracy also drops on held-out regions~\cite{gordon2026mmearth,doerksen2026earthshift,marsocci2024pangaea}.
Their cross-region evaluations, however, use single-date images, and most measure the drop after fine-tuning the encoder or training a deep decoder on its features~\cite{gordon2026mmearth,marsocci2024pangaea}. The drop of the pre-trained representation is therefore not isolated, and this setting differs from real-world application where users only train a lightweight classifier on frozen embeddings.
Closer to our realistic use of EO FMs, the EarthShift benchmark evaluates single-date frozen encoders with a lightweight convolutional decoder~\cite{doerksen2026earthshift} and finds a large drop under geographic shift with no additional robustness of EO FMs compared to generic vision models or fully supervised models.
For SITS FMs, the spatial generalization of frozen embeddings has been evaluated with shallow classifiers for crop types, but within a single region~\cite{lisaius2026towards}.
Overall, for both kinds of FMs, the drop is mostly measured on few transfer settings, in one direction, and reported as an aggregated score~\cite{gordon2026mmearth,doerksen2026earthshift,marsocci2024pangaea}.

% D. Pour ce faire on a besoin de datasets multi-region avec SITS
% il faut dire les jeux existants pour le geographical shift c'est XX, XX. TimeMatch est une proposition mais les labels disponibles sont au niveau pixel/parcelle" -> nothing written yet on TimeMatch / sykas as datasets.
Such evaluations are constrained by labeled data, since measuring spatial shift requires labels from different regions under a unified label taxonomy.
%  1. CROPS: timematch, eurocorps> rapidcrops, , sykas2022sentinel
For crop types, farmers' declarations have been harmonized across European countries~\cite{schneider2023eurocrops,rapidcrops}.
Multi-region SITS datasets therefore exist for crop types~\cite{nyborg2022timematch,sykas2022sentinel,reuss2025eurocropsml}.
TimeMatch~\cite{nyborg2022timematch}, for instance, covers four regions in three countries, with parcel-level labels.
%  2. TREES: one country, one nomenclature: TSA, grabska, pureforest, TROP LONG
For trees, to our knowledge, no such harmonized dataset exists over Europe. 
%is derived from forest-administration records in Germany, and the dataset of Grabska-Szwagrzyk et al.
%is derived from the Polish Forest Data Bank. 
% provides annotations of mono-specific stands in France
Each national dataset follows its own conventions, as for TreeSatAI in Germany~\cite{ahlswede2022treesatai}, the dataset of Grabska-Szwagrzyk et al.~\cite{grabska2024map} in Poland, or PureForest in France~\cite{gaydon2025pureforest}.
These datasets differ in label classes, types, geometry, and in acquisition years, as well as in how annotations are defined: they can describe either pure stands or dominant species of mixed stands.

% GAP
% linear probing eval: standard protocol to evaluate frozen representations, va avec operational use `feng2025tessera`, `brown2025alphaearth` + isole robustness des embedding comme ca le linear layer peut pas compenser (en justifiant qu'ils le font pas) -> on fait ca
% justiifer le probing sur frozen 
%  per-pair/per-class analysis instead of aggregated score -> on fait ca
% No multi-domain tree benchmark with unified label -> SPAN et RapidCrops
% > Si tu as dit ça plus haut [jeux existants + labels TimeMatch] on comprend pourquoi tu parles de extend TimeMatch
% No harmonized tree species European dataset exists, which makes tree species maps difficult to compare or create across regions.
In summary, to our knowledge, spatial generalization drop has not been measured for SITS FMs in a setting representative of the end-user needs, namely with a frozen SITS encoder and a shallow downstream classifier.
Moreover, because existing evaluations report aggregated scores, it remains unclear which domain pairs, transfer directions, or classes account for the drop.
Finally, harmonized multi-region labels exist for crop types, mostly at the parcel level, but not for tree species.
To address these gaps, we introduce \specdataname, a multi-domain benchmark that harmonizes the three tree species datasets presented above~\cite{grabska2024map,gaydon2025pureforest,ahlswede2022treesatai}, and \cropdataname, which provides pixel-level harmonized crop labels on the four TimeMatch tiles~\cite{nyborg2021timematchDataset,rapidcrops}. Using these benchmarks, we evaluate frozen SITS FMs through linear probing and report fine-grained results by domain pair, transfer direction, and class.

\subsection{Addressing the spatial generalization drop}
\label{sec:rel_addressing}
%~\cite{tseng2023lightweight} %must generalize to a wide range of geographies and tasks
%~\cite{feng2025tessera} %To ensure successful generalization, a multilevel data sampling, shuffling, and batching strategy was developed.
Two strategies can be employed to mitigate the performance drop when a model is applied to unseen regions.
The first adapts the model to each target region with domain adaptation techniques~\cite{nyborg2022timematch,li2024get,marsocci_2023_CVPR,gordon2026mmearth}. 
The second learns representations that transfer to new regions without any data from the target region, either during pre-training or when training the model on the downstream task.
We only detail the second strategy, as the first requires target data and per-region training, which does not fit real-world applications where the FM is pre-trained once and kept frozen.
Methods that follow this strategy act on three levers: the pre-training data, the training objective, and auxiliary variables.
%The first strategy does not fit our setting, in which the encoder is pre-trained once and kept frozen (\cref{sec:intro}).
%This section therefore reviews works that follow the second strategy.

%Data
A first lever is the pre-training data: sampling it across diverse regions has been motivated by spatial generalization in both SITS~\cite{tseng2023lightweight} and single-date FMs~\cite{danish2026terrafm}.
However, this does not appear sufficient on its own for spatial generalization: FMs pre-trained on datasets with global coverage are not more robust than models pre-trained on natural images or trained without pre-training  when they are tested on unseen regions~\cite{doerksen2026earthshift}.
%Studies that isolate the effect of the spatial distribution of pre-training data train and test the downstream classifier within the same region~\cite{purohit2025does,kaur2026pretrain}, so that the classifier is not evaluated on regions without labels.

%Objectives
A second lever is the training objective. 
Domain generalization methods aim to remove or isolate region-specific information, for example by separating region-specific from region-invariant features~\cite{ghassemi2025geographical}.
However, they are supervised and require class labels, often from several regions, which prevents their use during self-supervised pre-training.
Each end user would therefore have to train the encoder with these objectives, whereas FMs are meant to be used frozen.
%Moreover, across geographic regions, such methods did not outperform the same model with a standard classification objective~\cite{al2025analysing}.
In contrast, the general objective of EO FMs is to learn representations that generalize across tasks, regions, and periods~\cite{tseng2023lightweight,dumeur2025paving,brown2025alphaearth}.
For instance, TESSERA~\cite{feng2025tessera} attributes its generalization across sensors, seasons, and regions to an objective that makes the embeddings of a location invariant to random subsets of its cloud-free observations. Since both subsets come from the same location, this objective pushes the embeddings to be less dependent on the acquisition dates, but it does not explicitly target differences between regions, such as shifted crop calendars.

%Aux
A third lever, which does not require class labels, is the use of auxiliary variables: they describe environmental conditions that differ between regions and could thus account for regional differences during pre-training.
In fully supervised SITS models, replacing the day of year with thermal time, computed from temperature, improved transfer to held-out regions, although the gain varied with the transfer direction~\cite{nyborg2022generalized,turkoglu2026ts}.
In FMs, auxiliary variables are already integrated as inputs, targets, or both~\cite{tseng2023lightweight,tseng2025galileo,brown2025alphaearth,nedungadi2024mmearth}.
With task-specific models, In-N-Out~\cite{xie2021innout} compared auxiliary variables used as inputs or as targets on held-out regions, for cropland binary classification in the US Midwest and land cover classification in Africa.
Their results show that, used as inputs, auxiliary variables improved accuracy in the training regions but could reduce it in held-out regions, since their relation with the labels can shift across regions.
Used as pre-training targets, they improved accuracy in held-out regions.

% Gap
%However, In-N-Out compared the two roles at different training stages and with task-specific models.
%Within an FM kept frozen, a linear classifier trained on embeddings that contain auxiliary information could exploit relations between this information and the labels that do not hold across regions.
Although promising, this comparison has, to our knowledge, not been extended to SITS FMs.
We thus compare auxiliary variables used during pre-training either as inputs (\modelnamein) or as targets (\modelnameout), and assess their effect on the spatial generalization drop with frozen encoders.
%\modelname and its two variants share the same pre-training data, sampled across European ecoregions following the pre-training data lever.
%It therefore remains unclear how each role affects spatial generalization when auxiliary variables are integrated into the pre-training of a SITS FM that is then kept frozen.

%% file: sections/tables/related_encoders.tex
\providecommand{\nt}[1]{\textcolor{orange!85!black}{#1}}
\providecommand{\ntm}[1]{{\setlength{\fboxsep}{0.5pt}\colorbox{orange!35}{#1}}}
\begin{table}[t]
\caption{Representative pre-trained encoders discussed in \cref{sec:rel_fm_aux}.
\textbf{SSL}: reconstruction (R), joint-embedding (JE), discriminative (D).
\textbf{Spatial}: pixel-wise (Pixel), factorized sequential/parallel (Seq./Par.), joint/divided space-time attention (Joint/Div.).
\textbf{Temporal}: reduced downstream (DS), during pre-training (PT), or by learnable queries (Q).
\textbf{Aux.}: auxiliary variables as inputs (In) or targets (Tgt).
\textbf{Inf.}: are they needed at inference?
$^\dagger$: baseline in this study. SD: single-date. -: not applicable.}

\label{tab:related_fm}
\centering
\scriptsize
\setlength{\tabcolsep}{2.8pt}
\begin{tabular}{@{}llllccl@{}}
\toprule
 & & & & \multicolumn{3}{c}{Aux.} \\
\cmidrule(l){5-7}
Encoder & SSL & Spatial & Temporal & In & Tgt & Inf. \\
\midrule
TESSERA$^\dagger$~\cite{feng2025tessera}       & JE         & Pixel & PT  & \xmark & \xmark & - \\
ALISE$^\dagger$~\cite{dumeur2025paving}        & R+D        & Seq.  & Q   & \xmark & \xmark & - \\
AnySat$^\dagger$~\cite{astruc2025anysat}       & JE+D       & Seq.  & PT  & \xmark & \xmark & - \\
\greymidrule
Presto$^\dagger$~\cite{tseng2023lightweight}  & R          & Pixel & DS  & \cmark & \cmark & Optional \\
Galileo~\cite{tseng2025galileo}                & JE+D       & Joint & DS  & \cmark & \cmark & Optional \\
AlphaEarth~\cite{brown2025alphaearth}          & R+D        & Par.  & PT  & \xmark & \cmark & No \\
MMEarth (SD)~\cite{nedungadi2024mmearth}       & R     & -    & -  & \xmark & \cmark & No \\
\midrule
\modelname                                   & R          & Div.  & Q   & \xmark & \xmark & - \\
\modelnamein                                   & R          & Div.  & Q   & \cmark & \xmark & Required \\
\modelnameout                                  & R          & Div.  & Q   & \xmark & \cmark & No \\
\bottomrule
\end{tabular}
\end{table}

%% file: sections/tables/related_gen.tex
\providecommand{\nt}[1]{\textcolor{orange!85!black}{#1}}
\providecommand{\ntm}[1]{{\setlength{\fboxsep}{0.5pt}\colorbox{orange!35}{#1}}}
\providecommand{\pmark}{$\sim$}
\begin{table}[t]
\caption{Studies of the spatial generalization drop (\cref{sec:rel_eval}).
\textbf{SITS}: uses SITS?
\textbf{FM}: is a foundation model?
\textbf{Frozen}: is the model used frozen with only a shallow head trained?
\textbf{Held-out}: is it tested on unseen regions?
\textbf{Per pair/class}: are results reported per source-target pair/per class?
\textbf{Task}: task of the spatial evaluation.
\pmark{} means partly, - not applicable.}
\label{tab:related_spatgen}
\centering
\scriptsize
\setlength{\tabcolsep}{2pt}
\begin{tabular}{@{}lcccccc>{\raggedright\arraybackslash}p{1.75cm}@{}}
\toprule
Works & SITS & FM & Frozen & \makecell{Held-\\out} & \makecell{Per\\pair} & \makecell{Per\\class} & Task \\
\midrule
Orynbaikyzy et al.~\cite{orynbaikyzy2022spatial} & \cmark       & \xmark & -           & \cmark & \pmark & \cmark & Crop types \\
Sen4AgriNet~\cite{sykas2022sentinel}             & \cmark       & \xmark & -         & \cmark & \cmark & \cmark & Crop types\\
TimeMatch~\cite{nyborg2022timematch}             & \cmark       & \xmark & -           & \cmark & \cmark & \xmark       & Crop types \\
In-N-Out~\cite{xie2021innout}                    & \cmark     & \xmark & -           & \cmark & \xmark       & \xmark       & LC, cropland \\
\greymidrule
MMEarth-Bench~\cite{gordon2026mmearth}           & \xmark       & \cmark & \pmark & \cmark & \xmark       & \xmark       & Biomass, soil properties, mammal species\\
PANGAEA~\cite{marsocci2024pangaea}               & \xmark       & \cmark &  \xmark       & \cmark & \xmark       & \xmark       & Land cover \\
EarthShift~\cite{doerksen2026earthshift}         & \xmark       & \cmark & \cmark       & \cmark & \xmark       & \xmark       & Field boundaries \\
Lisaius et al.~\cite{lisaius2026towards}         & \cmark       & \cmark & \cmark       & \xmark & -           & -           & Crop types\\
\midrule
Ours                                      & \cmark       & \cmark & \cmark       & \cmark & \cmark       & \cmark       & Tree species, crop types \\
\bottomrule
\end{tabular}
\end{table}

%% file: sections/3_method.tex
\label{sec:method}
In this section, we first describe the \modelname architecture in~\cref{subsec:model}, before presenting its masked autoencoder pre-training scheme in~\cref{subsec:pre-training}. Finally, we detail two strategies to integrate auxiliary variables, either as inputs or as targets, respectively denoted \modelnamein and \modelnameout (\cref{subsec:aux-integration}).

\subsection{\modelname architecture} \label{subsec:model} 

\begin{figure}[t]
    \centering
    \includegraphics[width=0.95\columnwidth]{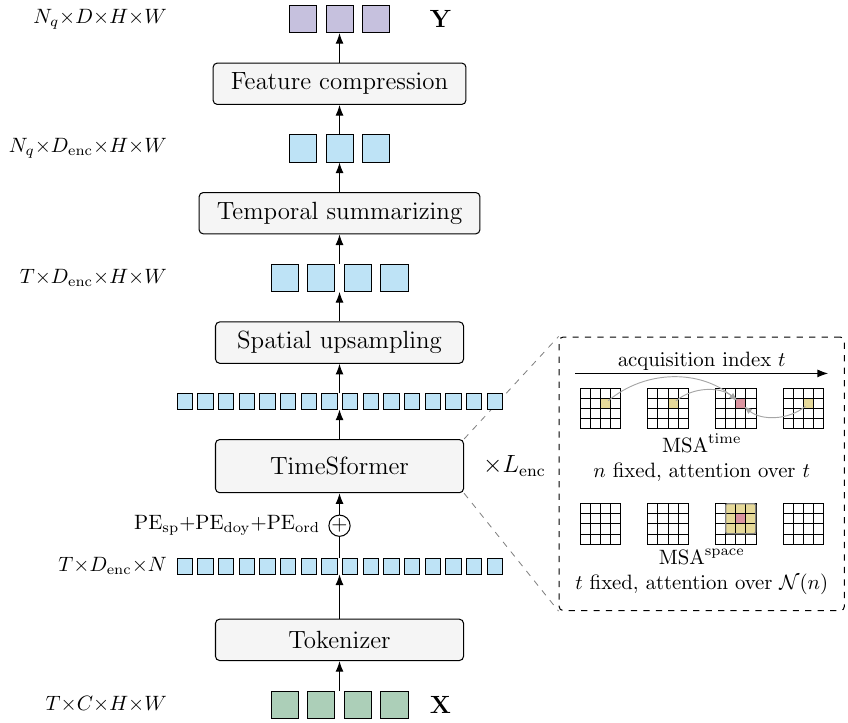}
    \caption{Architecture of the \modelname\ encoder, mapping an input SITS $\mathbf{X}$ to the latent representation $\mathbf{Y}$. The encoder consists of a patch tokenization with positional encodings, $L_{\mathrm{enc}}$ TimeSformer blocks, a spatial up-sampling, a temporal summarization with $N_q$ learnable queries, and a final MLP projection. The zoom details one TimeSformer block: temporal attention ($\mathrm{MSA}^{\mathrm{time}}$) over the acquisitions $t$ at a fixed spatial position $n$, followed by spatial attention ($\mathrm{MSA}^{\mathrm{space}}$) over the $3\times3$ token neighborhood
$\mathcal{N}(n)$ at a fixed acquisition $t$. The query token is shown in red and the attended tokens in yellow. Colors: green: input SITS; light blue: intermediate features; purple: fixed-size representations.}
    \label{fig:backbone}
\end{figure}

% general
% general
\modelname is a spatio-temporal encoder that maps SITS of variable length to a latent representation with a fixed temporal dimension, while preserving the spatial resolution of the input.

Formally, given an input SITS $\mathbf{X}$, \modelname outputs a latent representation $\mathbf{Y}$:
\begin{equation}
\mathbf{Y} = \modelnamemath(\mathbf{X}),  \,\, % because \quad to big
\modelnamemath : \mathbb{R}^{T \times C \times H \times W } \to \mathbb{R}^{N_q  \times D \times H \times W},
\end{equation} where $T$ denotes the number of acquisitions, which may vary from one series to another, $C$ the number of spectral channels, $H$ and $W$ the height and width of the input images, $N_q$ the fixed temporal dimension of the output, and $D$ the embedding dimension. We refer to the per-pixel vectors $\mathbf{Y}_{i,j}$ as embeddings.

% resume de tout
As illustrated in~\cref{fig:backbone}, \modelname consists of five main blocks: a patch tokenization stage producing a set of spatial tokens for each acquisition; a TimeSformer~\cite{bertasius2021space} backbone extracting spatio-temporal features; a spatial up-sampling of the tokens to the original input image resolution; a temporal summarization producing fixed-size latent representations; and a final projection reducing the embedding dimension.

% token + PE
First, each acquisition of the SITS is split into $N=HW/P^2$ patches of size $P \times P$, with $P=4$. The patches are projected to the model feature dimension $D_{\mathrm{enc}}=256$ by a linear layer $h$.
Three positional encodings ($\mathrm{PE}$) are then added to each resulting token: 
\begin{equation}
\mathbf{z}^{(0)}_{t,n} = h(\mathbf{x}_{t,n}) + \mathrm{PE}_{\mathrm{sp}}(n) + \mathrm{PE}_{\mathrm{doy}}(d_t) + \mathrm{PE}_{\mathrm{ord}}(t), 
\end{equation} where $h: \mathbb{R}^{P^2C}\to \mathbb{R}^{D_{\mathrm{enc}}}$, $\mathbf{x}_{t,n}$ is the flattened patch $n$ of acquisition $t$, and $d_t$ is the day of year (DOY) of acquisition $t$. $\mathrm{PE}_{\mathrm{sp}}$ is a fixed 2D sinusoidal encoding of the spatial position. $\mathrm{PE}_{\mathrm{doy}}$ is a sinusoidal temporal encoding based on the DOY~\cite{yuan2020self}. In addition, to allow the processing of multi-year time series, we add $\mathrm{PE}_{\mathrm{ord}}$, a learnable embedding of the index of each acquisition within the SITS. The tokenization thus produces a set of $T \times N$ tokens of dimension $D_{\mathrm{enc}}$, denoted $Z^{(0)}=\{\mathbf{z}_{1,1}^{(0)},\mathbf{z}_{1,2}^{(0)}, \ldots,\mathbf{z}_{T,N}^{(0)} \}$.

% backbone
Second, the tokens are processed by an adapted TimeSformer~\cite{bertasius2021space} backbone using divided space-time attention. It stacks $L_{\mathrm{enc}}=8$ blocks, where block $\ell$ maps $Z^{(\ell)}$ to $Z^{(\ell+1)}$ through a temporal attention layer followed by a spatial attention layer. In the temporal attention layer, each token $\mathbf{z}_{t,n}$ attends to the $T$ tokens at the same spatial position $n$. In the spatial attention layer, it attends to the tokens of its $3\times3$ spatial neighborhood $\mathcal{N}(n)$ within the same acquisition $t$. 
More precisely, we denote by $\mathrm{MSA}(\mathbf{q}, \mathcal{K})$ the multi-head self-attention (MSA) between the query $\mathbf{q}$ and the token set $\mathcal{K}$ (used as keys and values).
The two layers, illustrated in the zoom of~\cref{fig:backbone}, are written as:
\begin{align}
\mathbf{z}'^{(\ell)}_{t,n} &= \mathrm{MSA}^{\mathrm{time}}\big(\mathbf{z}^{(\ell)}_{t,n},\ \{\mathbf{z}^{(\ell)}_{t',n}\}_{t' \in \{1, \dots, T\}}\big),\label{eq:time_at} \\ 
\mathbf{z}^{(\ell+1)}_{t,n} &= \mathrm{MSA}^{\mathrm{space}}\big(\mathbf{z}'^{(\ell)}_{t,n},\ \{\mathbf{z}'^{(\ell)}_{t,n'}\}_{n'\in \mathcal{N}(n) }\big), \label{eq:space_at}
\end{align} where $\mathbf{z}'$ is the intermediate output of the temporal attention. Furthermore, as limited spatial modeling appears sufficient at Sentinel-2 resolution~\cite{garnot2019time} and the objects of interest are static, we slightly modify the spatial attention of the standard TimeSformer~\cite{bertasius2021space}. Specifically, we restrict the attention to the spatial neighborhood $\mathcal{N}(n)$~\cite{hassani2023neighborhood} (further detailed in \cref{app:local-attention}). Then, as in standard transformers~\cite{vaswani2017attention}, each attention layer is preceded by a layer normalization and wrapped in a residual connection, and each block except the last ends with a GEGLU feed-forward network~\cite{shazeer2020glu}.

% spatial up
Third, the tokens are up-sampled back to the original pixel resolution. The up-sampling module alternates bilinear interpolation and $3\times3$ convolutions, doubling the resolution at each stage until the input resolution is recovered. Up-sampling is applied independently to each acquisition, producing for each pixel $(i,j)$ a sequence of $T$ feature vectors $\mathbf{U}_{i,j} \in \mathbb{R}^{T \times D_{\mathrm{enc}}}$. 

% temp sum et 
Fourth, at each pixel, the temporal dimension is summarized with a cross-attention with $N_q=6$ learnable queries. We denote by $\mathrm{MCA}(Q, \mathcal{K})$ the multi-head cross-attention (MCA), where a set of queries $Q$, distinct from $\mathcal{K}$, attends to $\mathcal{K}$ (used as keys and values). The temporal summarization is then written as:
\begin{equation}
\bar{\mathbf{U}}_{i,j} = \mathrm{MCA}(Q, \mathbf{U}_{i,j}),
\end{equation} 
where $Q \in \mathbb{R}^{N_q \times D_{\mathrm{enc}}}$ are the learnable queries and $\bar{\mathbf{U}}_{i,j} \in \mathbb{R}^{N_q \times D_{\mathrm{enc}}}$ contains the $N_q$ temporal summary tokens of pixel $(i,j)$. The output temporal dimension is thus independent of $T$: the $N_q$ outputs are latent temporal features rather than per-acquisition features. 

% emb dim
Finally, each of the $N_q$ latent temporal features is projected from $D_{\mathrm{enc}}$ to $D=64$ by a two-layer multilayer perceptron (MLP) with hidden dimension 128, applied along the channel dimension. This produces the final representation $\mathbf{Y}_{i,j}=\mathrm{MLP}(\bar{\mathbf{U}}_{i,j}) \in \mathbb{R}^{N_q\times D}$.

\subsection{Pre-training} \label{subsec:pre-training}

\begin{figure}[t]
    \centering
    \includegraphics[width=\columnwidth]{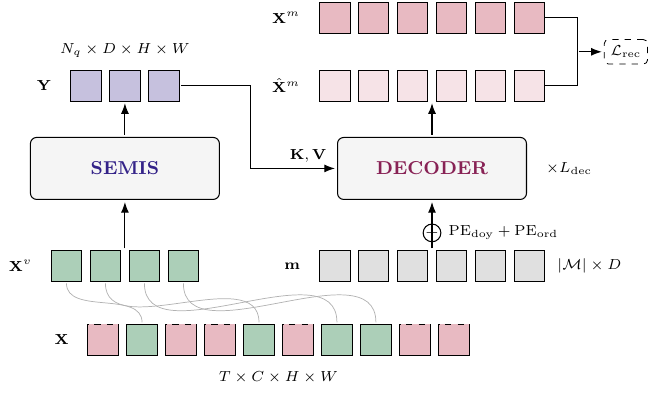}
   \caption{Self-supervised pre-training of \modelname. The acquisitions of the input SITS $\mathbf{X}$ are split into a visible sub-series $\mathbf{X}^v$, fed to the encoder, and a masked sub-series $\mathbf{X}^m$ (70\% of the acquisitions), whose dates are used to build the decoder queries. Each decoder block cross-attends to the latent representation $\mathbf{Y}$, used as keys and values ($\mathbf{K},\mathbf{V}$). The reconstruction $\hat{\mathbf{X}}^m$ is compared to the masked acquisitions $\mathbf{X}^m$ through the reconstruction loss $\mathcal{L}_{\mathrm{rec}}$, computed on non-cloudy pixels only.
   Colors as in \cref{fig:backbone}; in addition, dark pink: masked acquisitions $\mathbf{X}^m$; light pink: reconstructed acquisitions $\hat{\mathbf{X}}^m$; gray: learnable mask tokens $\mathbf{m}$.
   }
    \label{fig:pretraining}
\end{figure}

The pre-training of \modelname is a reconstruction task inspired by Masked Autoencoders (MAE)~\cite{he2022masked}. A dedicated decoder is trained to recover the masked acquisitions from the latent representation produced by the encoder. This stage is schematized in~\cref{fig:pretraining}. In this subsection, we first detail the masking strategy, then describe the decoder architecture, and finally present the reconstruction loss.

%The pre-training objective is designed to force the model to encode temporal structure rather than performing spatial inpainting.

% task
Our self-supervised pretext task consists of reconstructing masked acquisitions from a subset of the time series. Let $\mathcal{M}\subset\{1,\dots,T\}$ be the set of indices of the masked acquisitions, drawn uniformly at random with $|\mathcal{M}|=\lfloor 0.7\,T \rfloor$. This splits the input SITS into a masked sub-series $\mathbf{X}^{m}=\{\mathbf{X}_t\}_{t\in\mathcal{M}}$ and a visible sub-series $\mathbf{X}^{v}=\{\mathbf{X}_t\}_{t\notin\mathcal{M}}$. Only $\mathbf{X}^{v}$ is fed to the encoder. Masking is applied at the acquisition level: an acquisition is either kept or discarded in full, over all $H\times W$ pixels and $C$ channels. This design is intended to encourage the model to encode the temporal structure of the series, which reflects vegetation phenology and helps discriminate vegetation classes, rather than to perform spatial inpainting. 
As in MAE~\cite{he2022masked}, mask tokens are introduced in the decoder only: the decoder receives the dates of the masked acquisitions but never the corresponding images.
The encoder thus always processes actual acquisitions, which avoids a mismatch with downstream inputs, which contain no mask tokens.

Then, the decoder reconstructs the masked acquisitions from the latent representation of the visible sub-series.
The decoder operates at the pixel level and therefore extracts no spatial features.
It applies a succession of cross-attention blocks, then recovers the spectral channels through a linear projection.
More precisely, the initial query of a masked acquisition $t$ is the sum of a learnable mask token $\mathbf{m} \in \mathbb{R}^D$, shared across all masked acquisitions and pixels, and the same temporal positional encoding functions as in the encoder, computed at dimension $D$:
\begin{equation}
\mathbf{r}^{(0)}_t = \mathbf{m} + \mathrm{PE}_{\mathrm{doy}}(d_t) + \mathrm{PE}_{\mathrm{ord}}(t), \quad t \in \mathcal{M}.
\end{equation} The decoder then applies $L_{\mathrm{dec}}=2$ cross-attention blocks, in which the decoder queries attend to the encoder representation $\mathbf{Y}$:
\begin{equation}
 \mathbf{r}^{(\ell+1)}_{t,i,j} = \mathrm{MCA}\big(\mathbf{r}^{(\ell)}_{t,i,j},\, \mathbf{Y}_{i,j}\big), \quad \ell \in \{0, \dots, L_{\mathrm{dec}}-1\},
\end{equation}
where $\mathbf{r}^{(0)}_{t,i,j} = \mathbf{r}^{(0)}_t$ for all pixels $(i,j)$, and $\mathbf{Y}_{i,j} \in \mathbb{R}^{N_q \times D}$ provides the keys and values. The masked queries never attend to each other, so each masked acquisition is reconstructed independently at each pixel from $\mathbf{Y}_{i,j}$. Finally, the spectral values of each pixel are recovered through a linear projection:
\begin{equation}
    \hat{\mathbf{x}}_{t,i,j} = \phi(\mathbf{r}^{(L_{\mathrm{dec}})}_{t,i,j}),
\end{equation} where $\phi : \mathbb{R}^{D} \to \mathbb{R}^{C}$ is a linear layer applied independently at each pixel and each masked acquisition. We denote by $\hat{\mathbf{X}}_t \in \mathbb{R}^{C \times H \times W}$ the reconstructed acquisition $t$, and by $\hat{\mathbf{X}}^m = \{\hat{\mathbf{X}}_t\}_{t \in \mathcal{M}}$ the reconstructed masked sub-series.

% loss
The reconstruction loss is the mean squared error (MSE) between the masked acquisitions $\mathbf{X}^m$ and their reconstruction $\hat{\mathbf{X}}^m$, computed on non-cloudy pixels only:
\begin{equation}
\mathcal{L}_{\mathrm{rec}} = \frac{1}{N_{\mathrm{valid}}}
  \sum_{t \in \mathcal{M}}
  \big\| M^{\mathrm{clear}}_t \odot (\mathbf{X}_t - \hat{\mathbf{X}}_t) \big\|_2^2,
\label{eq:pt}
\end{equation} where $M^{\mathrm{clear}}_t \in \{0,1\}^{H \times W}$ is a binary mask indicating the non-cloudy pixels of acquisition $t$, broadcast over the $C$ channels, and $N_{\mathrm{valid}} = C\sum_{t\in\mathcal{M}}\|M^{\mathrm{clear}}_t\|_1$ is the total number of non-cloudy values over all masked acquisitions and channels.

\subsection{Auxiliary variable integration} \label{subsec:aux-integration}
\begin{figure}[t]
    \centering
    \subfloat[ \centering  \modelnamein  \label{fig:aux_in_int}]{%
        \includegraphics[width=0.49\linewidth]{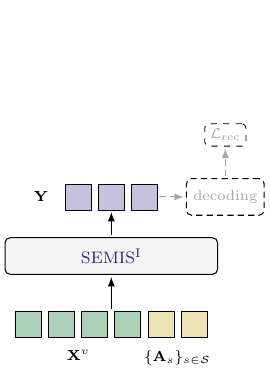}}
    \subfloat[ \centering \modelnameout \label{fig:aux_targ_int}]{%
       \includegraphics[width=0.49\linewidth]{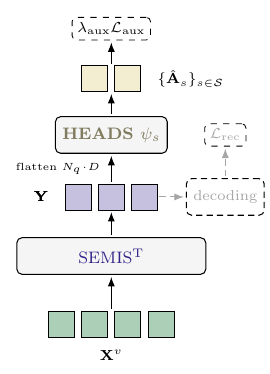}}
  \caption{Integration of auxiliary variables into the pre-training of \modelname. (a) \modelnamein: the auxiliary sources $\{\mathbf{A}_s\}_{s\in\mathcal{S}}$ are tokenized, appended to the visible acquisitions, and fed to the encoder as additional inputs. (b) \modelnameout: the auxiliary variables are predicted from the latent representation $\mathbf{Y}$ by source-specific heads $\psi_s$ and supervised by the auxiliary loss $\mathcal{L}_{\mathrm{aux}}$, weighted by $\lambda_{\mathrm{aux}}$. In both panels, the reconstruction branch, identical to that of~\cref{fig:pretraining}, is shown in abbreviated form (grayed and dashed).
    Colors as in \cref{fig:backbone,fig:pretraining}; in addition, yellow: auxiliary variables $\mathbf{A}_s$; lighter yellow: their predictions $\hat{\mathbf{A}}_s$.
  }
\end{figure}

We consider two ways of integrating auxiliary variables during the pre-training stage: either as additional inputs or as additional targets, denoted by the superscripts $\mathrm{I}$ and $\mathrm{T}$, respectively. The auxiliary variables themselves are described in \cref{subsec:aux}. Let $\mathcal{S}$ be the set of auxiliary sources used and $S=|\mathcal{S}|$ their count. The auxiliary data of source $s \in \mathcal{S}$ is denoted $\mathbf{A}_s \in \mathbb{R}^{ C_s \times H \times W }$, where $C_s$ is its number of variables. A model variant using source $s$ is denoted $\modelnamemath^{\mathrm{I}}_s$ or $\modelnamemath^{\mathrm{T}}_s$, depending on the integration strategy. In the following, \modelname without superscript refers to the model pre-trained on SITS only, without auxiliary variables.

\textbf{Auxiliary variables as inputs.} In \modelnamein, depicted in~\cref{fig:aux_in_int}, the $S$ auxiliary sources are appended to the input SITS as $S$ additional time steps.
Since the auxiliary variables are static and on the same spatial grid as the SITS, each source can be tokenized into the same $N$ spatial patches as an acquisition.
Appending these tokens along the temporal axis lets every acquisition token attend in each temporal attention layer to the auxiliary tokens at the same spatial position.
Each auxiliary source $s \in \mathcal{S}$, indexed from $1$ to $S$, is tokenized similarly to the acquisitions:
\begin{equation}
\mathbf{z}^{(0)}_{T+s,n} = h_s(\mathbf{a}_{s,n}) + \mathrm{PE}_{\mathrm{sp}}(n) + \mathrm{PE}_{\mathrm{src}}(s),    
\end{equation} where $h_s : \mathbb{R}^{P^2 C_s} \to \mathbb{R}^{D_{\mathrm{enc}}}$ is a source-specific linear projection, $\mathbf{a}_{s,n}$ is the flattened patch $n$ of $\mathbf{A}_s$, and $\mathrm{PE}_{\mathrm{src}}$ is a learnable embedding that identifies the source. The spatial encoding $\mathrm{PE}_{\mathrm{sp}}$ is the same as for the acquisition tokens. No temporal encoding is added, since the auxiliary sources employed are static. The concatenation of the image tokens and auxiliary tokens is fed to the backbone; the model thus operates over $T+S$ time steps.
During pre-training, the auxiliary tokens are never masked, and the loss $\mathcal{L}_{\mathrm{rec}}$ is unchanged.
For downstream tasks, the auxiliary tokens are also appended to the input SITS, which implies that they must be available over the target area at inference.
%We hypothesize that providing these variables as inputs makes the reconstruction task easier, as part of the information needed for reconstruction is given directly. For downstream tasks, this additional information is expected to improve in-domain performance compared to using spectral data alone. However, as discussed in~\cref{sec:related}, such inputs can also harm cross-domain transfer.

\textbf{Auxiliary variables as targets.} In \modelnameout, the auxiliary variables are used as additional pre-training targets (\cref{fig:aux_targ_int}). The model thus has to infer them from the input SITS alone. % This is expected to yield richer representations because the model has to deduce the environmental information from the SITS input. % expect the resulting features to be more comparable across regions
To this end, each auxiliary source $s \in \mathcal{S}$ is predicted from the latent representation by a dedicated head $\psi_s$:
\begin{equation}
\hat{\mathbf{a}}_{s,i,j} = \psi_s(\mathbf{Y}_{i,j}),
\end{equation} where $\psi_s : \mathbb{R}^{N_q D} \to \mathbb{R}^{C_s}$ is a source-specific MLP applied independently at each pixel to $\mathbf{Y}_{i,j}$, flattened into a vector of dimension $N_q D$. We denote by $\hat{\mathbf{A}}_s \in \mathbb{R}^{C_s \times H \times W}$ the prediction of source $s$ over all pixels.

The auxiliary loss and the full pre-training objective are then:
\begin{align}
\mathcal{L}_{\mathrm{aux}} &= \sum_{s \in \mathcal{S}} \frac{1}{N^{\mathrm{aux}}_s}
    \big\| M^{\mathrm{valid}}_s \odot (\hat{\mathbf{A}}_s - \mathbf{A}_s) \big\|_2^2, \\
\mathcal{L}_{\mathrm{total}} &= \mathcal{L}_{\mathrm{rec}} + \lambda_{\mathrm{aux}} \mathcal{L}_{\mathrm{aux}},
\end{align} with $\lambda_{\mathrm{aux}}=0.1$, where $M^{\mathrm{valid}}_s \in \{0,1\}^{H \times W}$ is the validity mask of source $s$, broadcast over its $C_s$ variables, and $N^{\mathrm{aux}}_s = C_s\|M^{\mathrm{valid}}_s\|_1$ is its number of valid values. Unlike \modelnamein, \modelnameout requires no auxiliary data at inference.

% M^{\mathrm{aux}}_s \odot (
%  where $N^{\mathrm{aux}}_s = C_s\|M^{\mathrm{aux}}_s\|_1$ is the number of valid (non-null) pixels and $M^{\mathrm{aux}}_s \in \{0,1\}^{H \times W}$  is the validity mask of the source.
%, some auxiliary have e.g SoiLGrid have NaN values in the surroundings of the cities.
% adds this aggregated auxiliary loss to the SITS reconstruction loss, weighted by $\lambda_\mathrm{aux}$ (\cref{eq:loss_total}). 

%% file: sections/4_data.tex
\label{sec:data}
In this section, we first describe the Sentinel-2 inputs shared by all datasets in \cref{subsec:s2}.
We then present the downstream datasets in  \cref{subsec:downstream} and the pre-training dataset in  \cref{subsec:pt}.
Finally, we detail the auxiliary variables in \cref{subsec:aux}.

\subsection{Sentinel-2 SITS inputs} 
\label{subsec:s2}
All datasets use Sentinel-2 (S2) Level-2A (L2A) SITS~\cite{sentinel2}\footnote{Sentinel-2 images were accessed from the AWS registry of open data at https://registry.opendata.aws/sentinel-2-l2a-cogs/.}.
We use the ten bands B02-B08, B8A, B11, and B12 ($C=10$), and the 20\,m bands are up-sampled to 10\,m using bilinear interpolation.
To limit the pre-processing cost, we download the three least cloudy images per month over the years covered by each dataset.
These images form the set from which the input series are sampled in the experiments (\cref{sec:xp}).
Cloud masks are generated with the deep-learning-based cloud detector of the CloudSEN12 project~\cite{aybar2022cloudsen12}, and provide the mask $M^{\mathrm{clear}}_t$.
S2 reflectances are normalized per band with global quantile statistics computed once over the pre-training dataset: values are clipped to the 0.5 and 99.5 percentiles, centered on the band median, and divided by the inter-quantile range.
The same statistics are applied to every downstream input using \modelname, so that no target-domain statistics are used at any stage.

\subsection{Downstream datasets}\label{subsec:downstream} 
Each task is associated with a \emph{multi-domain dataset}: \specdataname (tree species mapping, 3 domains) and \cropdataname (crop type mapping, 4 domains). In this paper, a \emph{domain} denotes a geographically distinct region with its own labeled data. In practice, this notion differs slightly between the two tasks. For \specdataname, the annotations of each domain come from a different tree species dataset, while the \cropdataname domains are different regions within the same annotation dataset (RapidCrops).

% , following TimeMatch's selection 

\begin{figure}
    \centering
      \includegraphics[width=\linewidth]{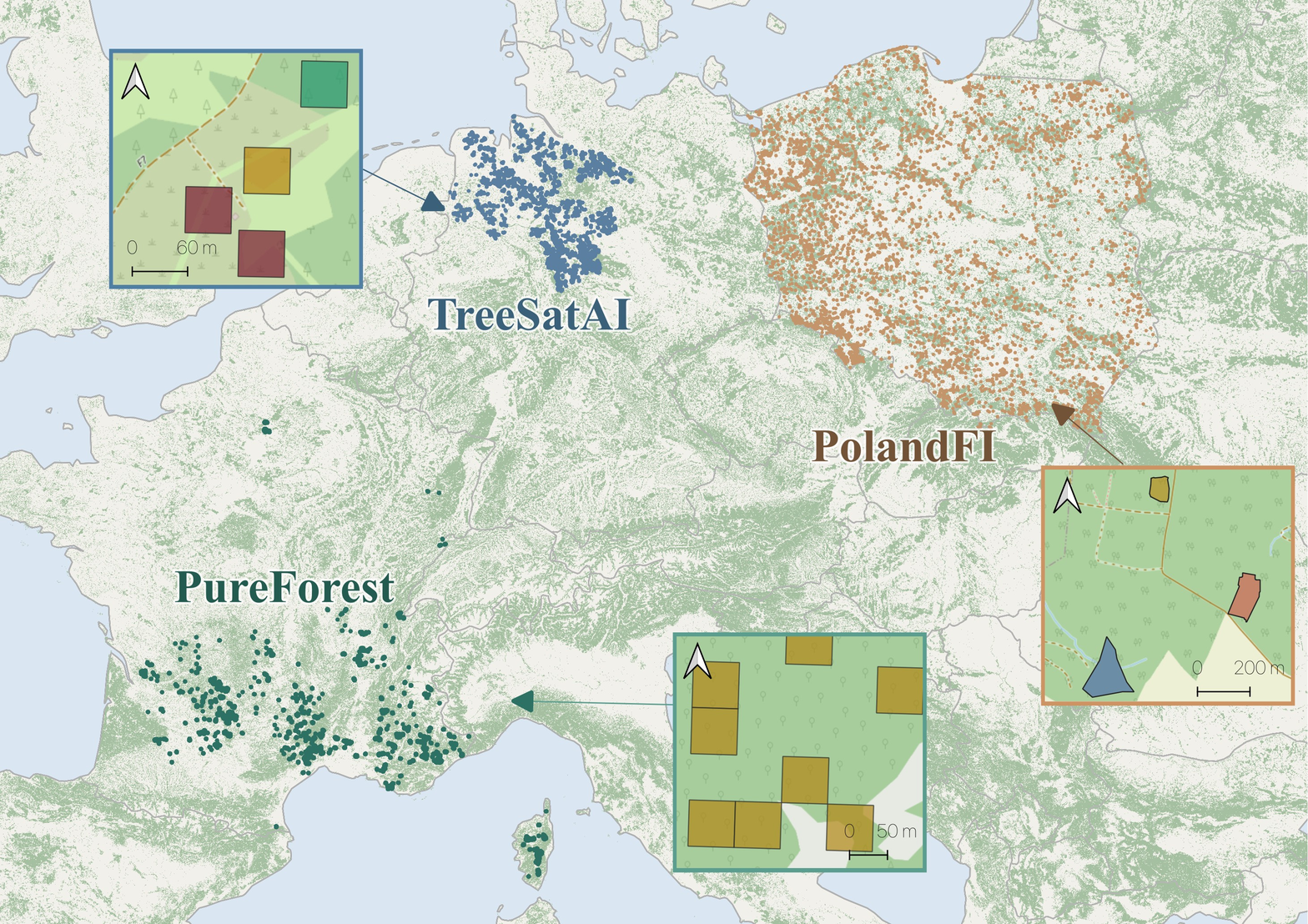}
     \caption{Geographic extent of the \specdataname domains.
    The zoom boxes show an example of annotations in each domain.
    Background: forest cover in 2020 from the JRC Global Forest Cover map (GFC2020, version 4)~\cite{bourgoin2026gfc2020}, land and country borders from Natural Earth. Inset backgrounds: \copyright~OpenStreetMap contributors~\cite{osm}.}
    \label{fig:coverage_sp}
\end{figure}

\begin{figure}
    \centering
     \includegraphics[width=\linewidth]{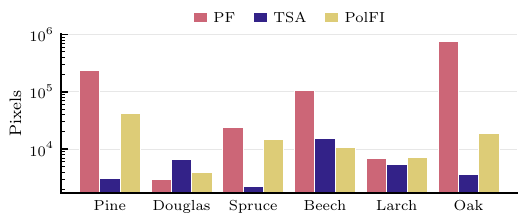}
    \vspace{-1em}
  \caption{Class distribution of \specdataname: number of labeled pixels per class and per domain (log scale).}
    \label{fig:support_sp}    
\end{figure}

\begin{table}
\caption{Description of the \specdataname domains.
\#Pixels: number of labeled pixels retained after harmonization into the six shared classes.
\#Classes: number of classes before harmonization.
Temp. Cov.: temporal coverage.}
\label{tab:species}
\setlength{\tabcolsep}{1.5pt}  
\centering
\begin{tabular}{l|cccccc}
\toprule
Name & Area & Temp. Cov. & \#Classes & \#Pixels & Type & Label Size \\
\midrule
PF &  France & 2024 & 13 &~1.1M & Square & $50\times50$m\\
PolFI  & Poland  & 2018-2021 &16& ~98K  & Polygons & - \\
TSA & Germany & 2017 &20 &~37K   & Square & $60\times60$m \\
\bottomrule

\end{tabular}
\end{table}

\subsubsection{Tree species multi-domain dataset}

We select three publicly available tree species datasets, spanning France, Germany, and Poland.
The aggregation of the three harmonized datasets is called \specdataname (tree SPecies Across Nations).
\cref{fig:coverage_sp} depicts their geographical coverage and an example of annotations for each domain.
As \cref{tab:species} shows, the three domains differ in every characteristic: region, acquisition period, native class granularity, label geometry, and number of labeled pixels.

Since no attribute is held constant across domains, cross-domain performance may not solely reflect geographic discrepancies.
Each dataset is released with its own original nomenclature, as its creators already grouped the raw taxa (from field inventories or photo-interpretation) into a reduced set of classes, at either the genus or the species level.
The class counts in \cref{tab:species} refer to these original classes, before the harmonization that allows the evaluation of the geographical domain shift.
Below, we first describe, for each dataset, the raw source and the mapping applied by its authors, then present our harmonization process.

\textbf{PureForest} (PF)~\cite{gaydon2025pureforest} is a mono-specific forest dataset: each patch is assigned to a single species.
Its annotations derive from the BD Forêt database~\cite{bdforetv2}, produced by photo-interpretation and cross-checked against National Forest Inventory field data~\cite{ign_ifn_data}.
The dataset mainly covers southern France.
Its authors grouped the 18 species into 13 classes.
We use the available train/validation/test splits, which ensure that no forest is shared across splits.

\textbf{TreeSatAI} (TSA)~\cite{ahlswede2022treesatai} is built from mixed forest-administration records in Lower Saxony, Germany~\cite{nlf2021a,nlf2021b}: permanent field-inventory plots and management polygons interpreted from aerial imagery.
Its authors aggregated these annotations into 20 classes covering 15 genera.
Following Astruc et al.~\cite{astruc2024omnisat}, we use the 2017 acquisitions and the splits defined in their work.

\textbf{PolandFI} (PolFI)~\cite{grabska2024map} is sampled from the Polish Forest Data Bank~\cite{bdl_poland}. It covers only public mixed-species forests and records, for each polygon, the shares of 29 tree species.
Its authors kept the majority species of each stand and mapped these species onto 16 classes.
We use the available train/test splits.
We derive a validation split from 10\% of the training data, using stratified random sampling to preserve the class distribution across splits.

Across the three datasets, the reference annotations are provided at different taxonomic levels, either genus or species.
Tree taxonomy is hierarchical: a genus groups several closely related species.
In addition, in some cases, a genus is split according to leaf phenology into two groups with different temporal patterns, rather than into species: deciduous trees, which shed their leaves seasonally, and evergreen trees, which keep their foliage all year.
Combining these annotation sources into a single multi-domain dataset therefore requires harmonizing their labels.
To build the \specdataname multi-domain dataset, we harmonize the three datasets into six classes that are available in all three datasets, either as explicit labels or derived from species-level annotations: Pine (genus \textit{Pinus}), Spruce (genus \textit{Picea}), Douglas (species \textit{Pseudotsuga menziesii}), Beech (genus \textit{Fagus}), Larch (genus \textit{Larix}), and (deciduous) Oak (genus \textit{Quercus}).
Pixels belonging to other classes are discarded.
Further details on the harmonization process are provided in \Cref{app:harmonization}.
\Cref{fig:support_sp} shows the label distribution across domains after harmonization.

Despite this harmonization, the domains still differ in label purity, i.e., the proportion of a labeled patch actually covered by the labeled class. As described above, PF describes patches that are strictly mono-specific forests, whereas TSA and PolFI labels are derived from forest inventories and correspond to the dominant species of each stand.

%First, mapping three distinct taxonomies into six classes necessarily discards finer species distinctions % and may increase intra-class variability unevenly across domains.

\subsubsection{Crop type multi-domain dataset} \label{subsec:crops}

\begin{figure}
    \centering
      \includegraphics[width=\linewidth]{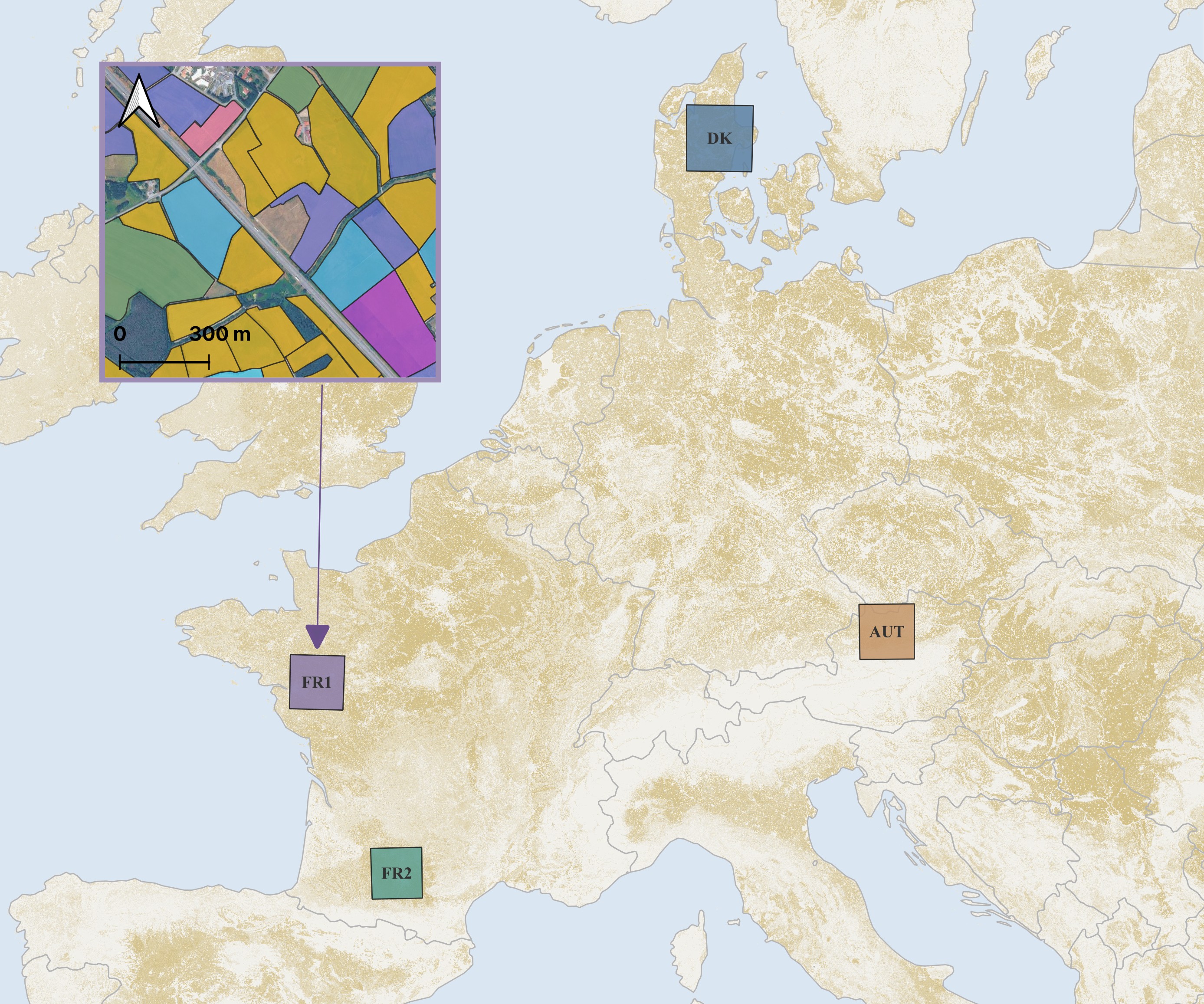}
\caption{Geographic extent of the \cropdataname domains.
The zoom box shows an example of annotations for FR1.
Background: crop cover fraction in 2019 from the Copernicus Global Land Cover map~\cite{buchhorn2020cgls}, land and country borders from Natural Earth.}
    \label{fig:coverage_crops}
\end{figure}

\begin{figure}
    \centering
     \includegraphics[width=\linewidth]{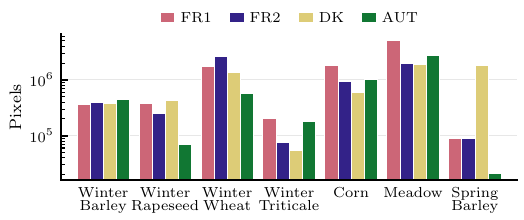}
     \vspace{-1em}
    \caption{Class distribution of \cropdataname: number of labeled pixels per class and per domain (log scale).}
    \label{fig:support_tm}
\end{figure}

\begin{table}
\caption{Description of the \cropdataname domains.
MGRS: Sentinel-2 tile identifier.
\#Pixels: number of labeled pixels retained after harmonization into the seven shared classes.}

\label{tab:crops}
\centering
\begin{tabular}{l|cccccc}
\toprule
Name & Area & MGRS & \# Pixels \\
\midrule
FR1 & France & 30TXT  & ~9.7M  \\
FR2 & France & 31TCJ  & ~6.4M  \\
DK  & Denmark  & 32VNH  & ~6.5M \\
AUT  & Austria  & 33UVP  & ~5.0M   \\
\bottomrule
\end{tabular}

\end{table}

We additionally evaluate our protocol on a crop type mapping task, using a dataset that we name \cropdataname (RapidCrops Spatial Generalization).
Its labels cover four domains in three European countries (\cref{fig:coverage_crops}).
\cref{tab:crops} summarizes the attributes of the domains, and \cref{fig:support_tm} shows the class distribution after harmonization.

\cropdataname is built from RapidCrops~\cite{rapidcrops}, a polygon-based crop type dataset whose labels follow the HCAT3 taxonomy~\cite{schneider2023eurocrops}, a harmonized crop class definition across European farmers' declarations.
To enable spatial cross-domain evaluation, we adopt the four S2 tiles of the Military Grid Reference System (MGRS) selected in the TimeMatch dataset~\cite{nyborg2021timematchDataset}: 32VNH (DK), 30TXT (FR1), 31TCJ (FR2), and 33UVP (AUT).
These tiles cover three European countries with known phenological shifts between regions.
Unlike the existing TimeMatch dataset, which provides parcel-level annotations, our dataset (\cropdataname) provides labels at the pixel level.
Regarding the annotations, we preserve only the classes that are present in all four domains and represent at least 0.4\% of the samples in each. The resulting classes are Corn, Meadow, Spring Barley, Winter Barley, Winter Rapeseed, Winter Triticale, and Winter Wheat. 
\cropdataname consists of SITS and raster labels whose spatial and temporal extents are defined as follows.
MGRS tiles are divided into $64 \times 64$ pixel patches, and only patches containing at least one labeled parcel are kept.
Each patch is randomly assigned a single year (2020 or 2021), balanced within each $512 \times 512$ block, and both its labels and its corresponding S2 time series are restricted to that year.
Patches are then randomly split into training, validation, and test sets (70/15/15\%).

For both multi-domain datasets, all labels are rasterized onto the native 10\,m S2 grid. % with the same coordinate reference system (CRS), origin, and transform as the tile.
A pixel is labeled if any part of a geometry touches it, with no minimum-overlap threshold.
Unlabeled pixels are set to a no-data value.

\subsection{Pre-training dataset} \label{subsec:pt}

\begin{figure}
    \centering
    \includegraphics[width=\linewidth]{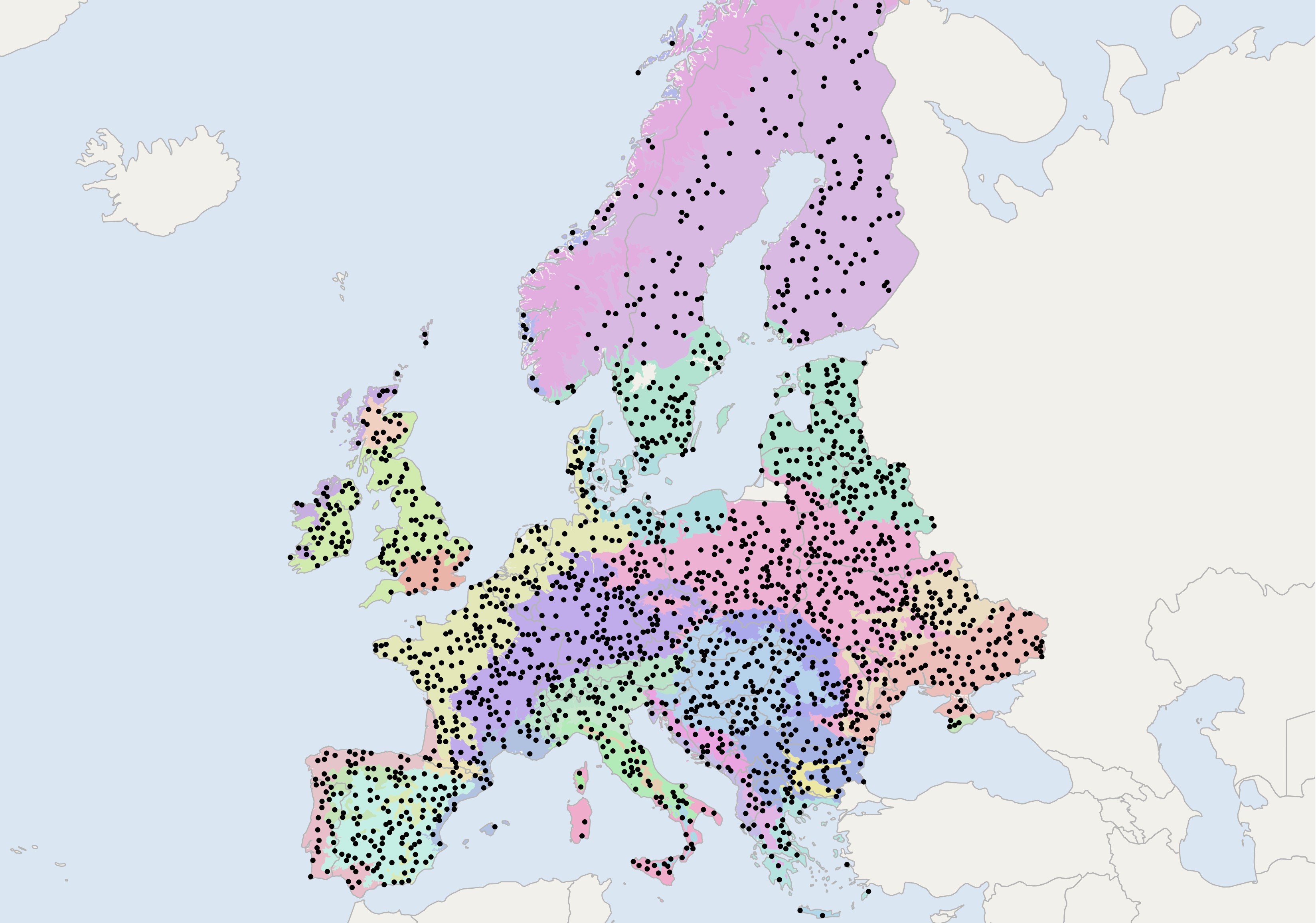}
     \caption{Coverage of the pre-training dataset. Black circles show the positions of the SITS, and colors show the ecoregions~\cite{dinerstein2017ecoregion}. Base map: Natural Earth.}
    \label{fig:coverage_pt}
\end{figure}

The pre-training dataset consists of 2000 S2 L2A sites of $512\times512$ pixels over Europe (\cref{fig:coverage_pt}), for a total of 2.6\,TB.
To promote diversity across the input SITS, the sites are sampled across European ecoregions~\cite{dinerstein2017ecoregion}.
In most cases, the number of sites allocated to each ecoregion is proportional to its area, with a minimum of one site per ecoregion.
The \textit{Scandinavian and Russian taiga} ecoregion is an exception: its allocation is down-weighted by a factor of 0.3 to avoid over-representation. This ecoregion covers a large area but has a relatively limited vegetation diversity, as it is dominated by only a few tree species~\cite{tikhonova2017tree}.

\subsection{Auxiliary variables} \label{subsec:aux}  

\begin{table}
  \caption{Description of the auxiliary data.
  $C_s$: number of variables of source $s$.
  FC: field capacity; PWP: permanent wilting point; TAWC: total available water content.}
    \label{tab:aux}
    \setlength{\tabcolsep}{3pt}
    \centering  
    \begin{tabular}{llcc}
    \toprule
        Name & Variables & $C_s$ & Sources \\
        \midrule 
      topo & slope, $\sin$(aspect), $\cos$(aspect) & 3 &~\cite{copernicus_dem,RichDEM} \\
      soil & pH, FC, PWP, TAWC & 4 &~\cite{poggio2021soilgrids,panagos2022european} \\ 
      era5 & \makecell[l]{T$_{\min}$, T$_{\max}$, precip., wind, dew point, \\ vapor pressure, radiation} & 7 &~\cite{boogaard2020agrometeorological} \\
      \bottomrule
    \end{tabular}
\end{table}

We use three sources of auxiliary variables: topographic (topo), soil properties (soil), and climatic (era5) variables.
Following the notation of \cref{subsec:aux-integration}, $s \in \{\text{topo}, \text{soil}, \text{era5}\}$, with $C_{\text{topo}}=3$, $C_{\text{soil}}=4$, and $C_{\text{era5}}=7$.
The variables of each source are summarized in \cref{tab:aux}.
In our experiments, we use either one source at a time ($S=1$) or the combination of the three sources ($S=3$), denoted as ``combo''. %$\modelnamemath^{\mathrm{T}}_{\mathrm{combo}}$, which is only evaluated for \modelnameout.

\textbf{Topographic} (topo) variables consist of the slope and aspect derived from the Copernicus GLO-30 Digital Elevation Model (DEM)~\cite{copernicus_dem}, computed with the RichDEM terrain analysis library~\cite{RichDEM}.
The aspect, i.e., the compass direction that a slope faces, is a circular variable and is therefore encoded by its sine and cosine. %Considering the two auxiliary integration proposed in ~\cref{subsec:aux-integration}, we posit that elevation cannot be predicted from multi-spectral reflectance alone. Thus, we exclude this information from the model using auxiliaries as targets \modelnameout as well as in the other variant, for fair comparison. 

%Thus we  Absolute elevation is excluded on the basis that it cannot be recovered from multi-spectral reflectance alone, which makes it an unsuitable reconstruction target in Phen-G.

\textbf{Soil} (soil) properties variables consist of the water pH, field capacity (FC), and permanent wilting point (PWP) at 15-30\,cm depth from the SoilGrids global maps~\cite{poggio2021soilgrids}, and of the total available water content (TAWC) at 30-100\,cm depth from the European Soil Database Derived data of the European Soil Data Centre~\cite{panagos2022european}.

%enabling better resolution and better quality information since it is at European scale (au cas ou demandé en review pk on a 2 sources pour le soil) 

\textbf{Climatic} (era5) variables consist of seven variables from AgERA5~\cite{boogaard2020agrometeorological}, an agrometeorological dataset derived from the ERA5 reanalysis: 24-hour minimum and maximum temperature, precipitation flux, vapor pressure, 10\,m wind speed, 2\,m dew point temperature, and solar radiation.
These variables describe climate conditions relevant to vegetation phenology.
Daily values are aggregated into monthly statistics: the mean for temperature, radiation, vapor pressure, dew point temperature, and wind speed, and the sum for precipitation.
Given the coarse spatial resolution of the climate data, climatic variables are nearly constant within an input patch; each patch is therefore assigned a single value per variable.
For both integration variants, these monthly values are then averaged over the months and years covered by each input SITS, producing one vector per patch that summarizes its climate conditions. 
This summary discards the temporal dimension of climate, as all auxiliary sources are encoded as static data (\cref{subsec:aux-integration}).

The topographic and soil variables are reprojected onto the 10\,m S2 grid (the SITS spatial resolution) using cubic-spline resampling.

%% file: sections/5_xp.tex
\label{sec:xp}
In this section, we first present the spatial generalization evaluation protocol in \cref{subsec:val_protocol}, designed to assess how the representations of frozen pre-trained models transfer across spatial domains.
We then detail, in~\cref{subsec:imp}, the pre-training settings, the baseline models used for comparison with \modelname, and the linear probing and fully supervised training settings.

\subsection{Spatial generalization evaluation protocol}
\label{subsec:val_protocol}
\begin{figure}[t]
    \centering
    \includegraphics[width=\linewidth]{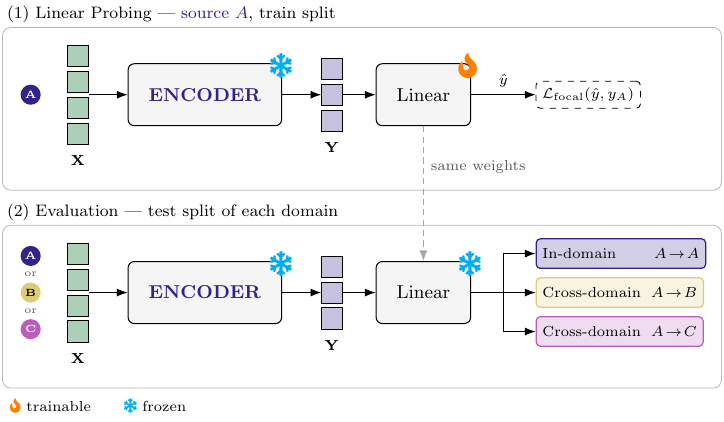}
    \caption{Spatial generalization evaluation protocol.
    (1) Linear probing: the pre-trained encoder is frozen, and only a linear layer is trained with a focal loss on the training split of a single source domain $A$.
    (2) Evaluation: the same frozen encoder and linear layer are applied without any adaptation to the test split of each domain.
This yields one in-domain macro F1 score (mF1) ($A$ \ra $A$) and one cross-domain mF1 per direction ($A$ \ra $B$, $A$ \ra $C$).
}%The procedure is repeated with each domain as the source.}
    \label{fig:protocol}
\end{figure}

To assess the spatial transferability of the representations produced by the different models, we follow an evaluation protocol similar to that of~\cite{nyborg2022timematch}, depicted in \cref{fig:protocol}.
We call \emph{in-domain} the setting in which a model is trained on a spatial region $A$ (the \emph{source} domain) and tested on the same region $A$ (the \emph{target} domain).
In contrast, in the \emph{cross-domain} setting, a model trained on region $A$ is tested on another region ($B$ or $C$).
We call \emph{direction} a source$\to$target pair, and \emph{transfer} the application of a model trained on the source domain to a different target domain, without any adaptation.

In practice, we apply this protocol to the downstream evaluation of frozen pre-trained backbones (baseline FMs or \modelname).
As shown in \cref{fig:protocol}, the pre-trained model remains frozen during training on the source domain, and only a linear classifier is trained on top of its embeddings.

%In our protocol, we chose to use a single source domain because with several sources, each transfer result would mix different spatial shifts, and a drop would be harder to attribute to a given source/target pair. 

%%%%%%%%%%%%%%%

\subsection{Implementation details} \label{subsec:imp}
\subsubsection{Pre-training}
We pre-train three variants of the model: \modelname, \modelnamein, and \modelnameout.
Each variant with auxiliary variables is trained with a single source (era5, soil, or topo).
\modelnameout is additionally trained with the combination of the three sources ($\modelnamemath^{\mathrm{T}}_{\mathrm{combo}}$).
The models are pre-trained for 400 epochs with the AdamW optimizer, a maximum learning rate of $10^{-4}$, and a OneCycleLR scheduler\footnote{\url{https://docs.pytorch.org/docs/2.12/generated/torch.optim.lr\_scheduler.OneCycleLR.html}} (details in~\Cref{app:hyperparameters}).
Training is performed on two H100 GPUs with mixed precision and a batch size of 32 per GPU (64 in total).
The model is pre-trained on $64\times64$ pixel patches obtained by regularly tiling each $512\times512$ site.
Each input SITS covers two randomly chosen consecutive years, with one acquisition per month ($T=24$).
At each epoch, this acquisition is randomly selected among the three least cloudy images downloaded for that month (\cref{subsec:s2}).
As described in~\cref{subsec:pre-training,subsec:aux-integration}, 70\% of the acquisitions are masked, and \modelnameout uses $\lambda_{\mathrm{aux}} = 0.1$.

\subsubsection{Baseline models}  \label{subsec:bench_models}
We select four encoders that natively process S2 SITS and output fixed-size embeddings at S2 spatial resolution. To ensure that all models receive the same S2 input, we also require that their pre-trained weights be publicly available, instead of relying on pre-computed embeddings. The four selected baseline models are: Presto\footnote{\url{https://github.com/nasaharvest/presto/tree/main}}, ALISE\footnote{\url{https://zenodo.org/records/10838982}}, AnySat\footnote{\url{https://github.com/gastruc/AnySat\#install-and-load-anysat}}, and TESSERA\footnote{\url{https://docs.torchgeo.org/en/latest/api/models/tessera.html}}.
All inputs are normalized with the original normalization method and statistics of each model.
Presto~\cite{tseng2023lightweight} is fed the S2 SITS together with Dynamic World land cover maps and coordinates.
Its embedding dimension is $128$ for each pixel.
ALISE~\cite{dumeur2025paving} is fed the S2 SITS.
Its output embedding has shape $H \times W \times 10 \times 64$, where $10$ is the number of learnable queries along the temporal dimension and $64$ the embedding dimension.
We concatenate the $10$ temporal features along the embedding dimension and obtain embeddings of size $640$.
AnySat~\cite{astruc2025anysat} is fed the same input, and we keep only its dense embeddings of dimension $768$ for each pixel.
Note that AnySat was pre-trained on the TreeSatAI inputs (without annotations), which could improve its performance on this domain.
For TESSERA~\cite{feng2025tessera}, we use only the branch dedicated to S2 SITS, which processes pixel-level time series.
We denote this version TESSERA-S2.
Its embedding dimension is $512$ for each pixel.

\subsubsection{Linear probing}\label{subsec:probing}
All models are evaluated with the same configuration: a linear layer is trained on top of the frozen encoder.
The input SITS consist of one randomly selected image per month over one year ($T=12$).
For \modelnamein, the $S$ auxiliary tokens are appended, giving $T+S$ time steps.
Similar to ALISE, the $N_q=6$ temporal features of \modelname and its variants are concatenated along the embedding dimension, which gives embeddings of size $N_q D = 384$.
The loss is a focal loss with $\gamma=2$.
We train the linear layer for 250 epochs with the AdamW optimizer, a learning rate of $10^{-3}$, and a CosineAnnealingLR scheduler\footnote{\url{https://docs.pytorch.org/docs/2.12/generated/torch.optim.lr\_scheduler.CosineAnnealingLR.html}}.
The weights with the best score on the validation split of the source domain are evaluated.
%This validation split is the only one used, so that no target data is seen at any point of training or model selection.
The linear layer is trained on two V100 GPUs with an effective batch size of 256 for all models, except for \modelname and its variants on \specdataname, which use 128.
For each patch, the monthly images are drawn once and kept fixed across epochs. %To reduce computation, we extract frozen encoder features once per patch before training the linear head, such that each encoder requires only a single forward pass per patch.

%We report macro-F1 (mF1) because vegetation species datasets are highly imbalanced reflecting their real-world imbalance. All reported scores come from a single run per configuration.

\subsubsection{Fully supervised baseline}\label{subsec:sup}
As a reference on downstream task performance, we also report the results of \modelnameFS, trained in a fully supervised setting.
This model uses the \modelname architecture, with a linear head on top of the embeddings concatenated along the embedding dimension, as in~\cref{subsec:probing}.
The encoder and the head are trained jointly from scratch, without pre-training.
The training hyperparameters (learning rate, scheduler, loss, and number of epochs) are the same as in the linear probing setting.

%% file: sections/6_results.tex
\label{sec:results}
%%%%%%%%%%%%%%%%%%%%%%%
In this section, we analyze the in-domain and cross-domain performance on the \specdataname and \cropdataname multi-domain benchmarks, following our spatial generalization evaluation protocol (\cref{sec:xp}).
First, we compare the in-domain and cross-domain results of the default \modelname configuration (frozen pre-trained encoder) with those of its fully supervised counterpart \modelnameFS, to assess the contribution of self-supervised pre-training (\cref{subsec:ssl_contrib}).
Second, we analyze the spatial generalization of \modelname and the baseline models (\cref{subsec:spatial_gen}): their overall performance, then their in-domain (\cref{subsec:benchin_dom}) and cross-domain (\cref{subsec:bench_cross}) results.
Finally, we compare the integration of auxiliary variables as inputs or as pre-training targets, and examine whether they mitigate the cross-domain patterns identified in \cref{subsec:bench_cross} (\cref{subsec:aux_res}).

\input{sections/tables/in_cross_sp}
\input{sections/tables/in_cross_tm}

% SSL contribution againt supervised 
\subsection{Contribution of self-supervised pre-training}\label{subsec:ssl_contrib}

We investigate whether a frozen encoder pre-trained with a standard MAE strategy can match or outperform a fully supervised model, in-domain and cross-domain.
To this end, we compare \modelname in a linear probing setting with \modelnameFS, its fully supervised counterpart presented in~\cref{subsec:sup}.
In the linear probing setting, the pre-trained encoder is kept frozen and only a linear layer is trained (\cref{subsec:probing}).
\Cref{tab:in_cross_sp,tab:in_cross_tm} report the results of the two models and their number of trainable parameters.
For each task, we report the macro F1 (mF1) scores on the test set of the target domains.
As defined in~\cref{sec:xp}, in-domain scores are obtained when evaluating the model on the source domain, and cross-domain scores are obtained when evaluating the same model on a different target domain.
The differences between the two models reported below are averaged over all target domains.

In the in-domain setting, \modelnameFS outperforms \modelname by 6.7 mF1 points for tree species and by 10.8 for crop types.
This difference may be partly explained by the smaller number of trainable parameters of \modelname, whose encoder is frozen, compared to \modelnameFS.
In the cross-domain setting, \modelname outperforms \modelnameFS by 9.4 mF1 points for tree species, despite having fewer trainable parameters.
This result suggests that the pre-trained representations transfer better across domains for this task.
For crop types, however, \modelnameFS keeps its advantage in the cross-domain setting with +7.2 mF1 compared to \modelname. %but all the baselines too.
It seems that, as detailed in \cref{subsec:crops}, crop labels are spatially denser (\cref{fig:coverage_crops}) and possibly more consistent across domains than tree species labels. % since their labels are based on a uniform identification format.
Thus, during training for crop-type mapping, the features learned by \modelnameFS on the source domain still appear relevant in the target domains. % since their labels are based on a uniform identification format. 
Conversely, the \specdataname benchmark contains fewer samples and sparser labels. Consequently, \modelnameFS may overfit to the source domain for tree species, which could limit its transfer to other domains.

Overall, despite far fewer trainable parameters, using \modelname, the difference between the in-domain and cross-domain scores is smaller on both tasks, showing that self-supervised pre-training mainly benefits spatial generalization, especially when labels are scarce.

%%%%%----- BENCH ----%%%
\subsection{Spatial generalization analysis}\label{subsec:spatial_gen}
We now characterize the spatial generalization of \modelname and the baseline models on the two tasks.
As in~\cref{subsec:ssl_contrib}, \cref{tab:in_cross_sp,tab:in_cross_tm} report their in-domain and cross-domain performance.
We first study the global performance of the benchmarked models. We then analyze their in-domain and cross-domain performance at the domain and class levels for each direction (source\ra target pair) with the cross-domain drop (in-domain F1 on the target domain minus cross-domain F1).

%%%%%%%%%%%%%%%%%%%%% PRES MODELS %%%%%%%%%%%%%%%%%%%%%
\vspace{0.4em}
\subsubsection{Global performance of the benchmarked models} In this section, we investigate the results of the different benchmarked models across tasks.

The three best models, \modelname, AnySat, and TESSERA-S2, obtain similar performance on both tasks, in-domain and cross-domain (within 7.6 mF1 points of each other,~\cref{tab:in_cross_sp,tab:in_cross_tm}).
In general, the ranking of the models follows their number of frozen parameters.
There are two exceptions to this rule.
First, \modelname outperforms AnySat in-domain on crop types.
Second, in the cross-domain setting, TESSERA-S2 outperforms AnySat on both tasks while having fewer parameters, although AnySat performs better in-domain.

From in-domain to cross-domain, the mF1 of these frozen encoders drops by 30 to 36 points on \specdataname and by 19 to 36 points on \cropdataname, where Presto has the largest drop~(\cref{tab:in_cross_sp,tab:in_cross_tm}).
Presto obtains lower scores than all the other models on both tasks: about 20 mF1 points below their average in-domain, and up to 35 points below their average cross-domain on crop types.
A first possible explanation for these lower scores is the small number of parameters of Presto.
We also hypothesize that its large cross-domain drop may be caused by the auxiliary inputs it uses during linear probing and inference (coordinates and Dynamic World land-cover layers).
These additional inputs may lead to very different representations across geographical regions.
%This hypothesis is consistent with the results of \modelnamein in~\cref{subsec:aux_res}.
Thus, to prevent Presto's scores from lowering the averaged metrics, we exclude it from the metrics averaged across baseline encoders in the following subsections.

%%%%%%%%%%%%%%%%%%%%% IN DOMAIN %%%%%%%%%%%%%%%%%%%%%
\vspace{0.4em}
\subsubsection{In-domain performance}\label{subsec:benchin_dom} 
We first analyze in-domain performance to contextualize the cross-domain results presented in~\cref{subsec:bench_cross}, where the drop of each class is measured against its in-domain score on the same target domain.
This drop is only meaningful for classes that are well classified in-domain: when a class is already poorly classified within a domain, low cross-domain scores cannot be attributed to the change of domain.
We therefore identify these difficult classes, and exclude them from both the in-domain and cross-domain averages and analysis when their domain is the target.
\Cref{tab:in_cross_sp,tab:in_cross_tm} still report scores on all classes.
%We consider a class difficult in a domain when its average F1 is below 50 or its best F1 is below 60 which means that no benchmarked model classify it well.

\begin{figure}
    \centering
    \includegraphics[width=0.6\linewidth]{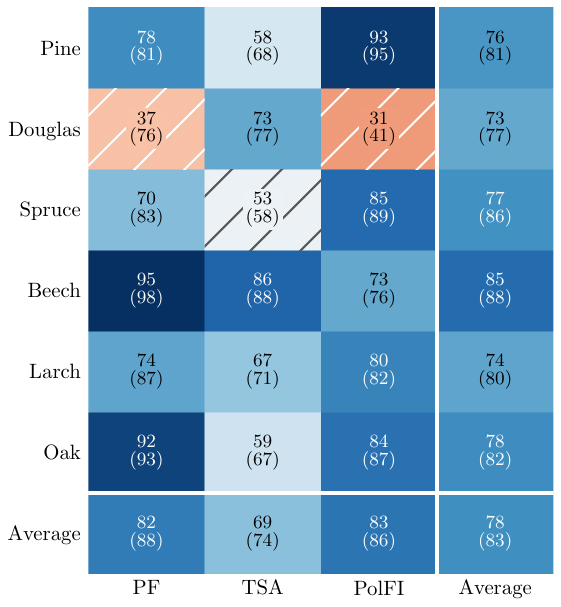}
    \caption{In-domain F1 per class and target domain on \specdataname, averaged over the benchmarked models (Presto excluded). The highest F1 among the models is given in brackets. The color scale is centered on 50. A low average F1 indicates that the class is difficult to classify in this domain across the benchmarked models. Difficult classes are hatched and excluded from the average row and column.}
    \label{fig:ind_f1_class_sp}
\end{figure}

\begin{figure}
    \centering
    \includegraphics[width=0.75\linewidth]{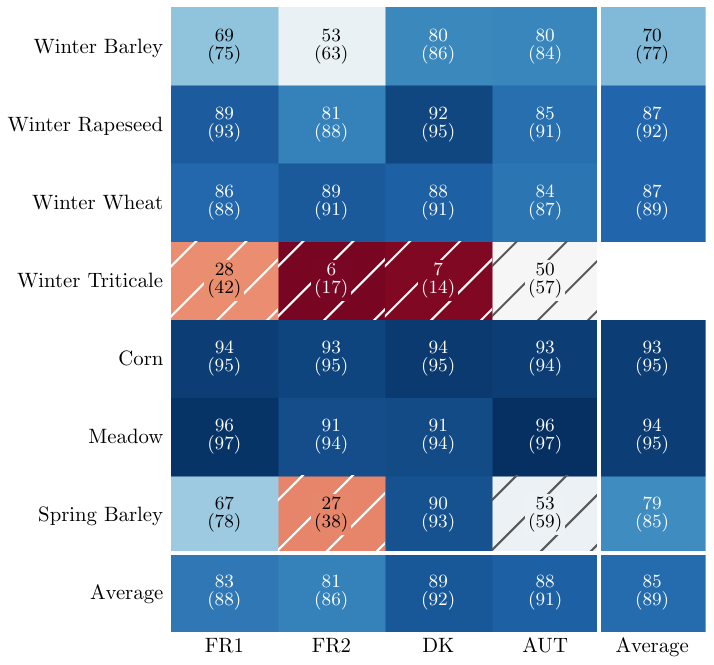}
        \caption{In-domain F1 per class and target domain on \cropdataname, averaged over the benchmarked models (Presto excluded). The highest F1 among the models is given in brackets. The color scale is centered on 50. A low average F1 indicates that the class is difficult to classify in this domain across the benchmarked models. Difficult classes are hatched and excluded from the average row and column.} 
    \label{fig:ind_f1_class_tm}
\end{figure}

\Cref{fig:ind_f1_class_sp,fig:ind_f1_class_tm} report the per-class, per-domain F1 scores, averaged across ALISE, AnySat, TESSERA-S2, and \modelname.
A low average F1 score indicates that the corresponding class is difficult to classify across these models within a domain.

Overall, \modelname and the baseline models obtain an average in-domain mF1 of 78 for tree species and 85 for crop types, once the difficult classes identified below are excluded (\cref{fig:ind_f1_class_sp,fig:ind_f1_class_tm}).
Performance nonetheless varies across tasks, domains, and classes, as detailed in the following analysis.

% species When investigating tree species mapping,
\textbf{In-domain tree species mapping.}
We observe two main trends in~\cref{fig:ind_f1_class_sp}.

First, Douglas fir is difficult to classify in the PF and PolFI domains, but not in TSA.
In both domains, it is mainly confused with other evergreen conifers, in particular Pine (confusion matrices in~\Cref{app:indomain}).
Douglas firs and Pines are evergreen conifers that share similar ecological niches.
Their physical differences, such as needle shape and arrangement, cannot be captured at a 10~m spatial resolution.
In addition, because they keep their needles year-round, they lack the strong seasonal variations that could help separate them.
As a result, their spectro-temporal signatures are likely very similar, which would explain the high confusion between them.
%Pine itself is only marginally affected by this confusion, as Douglas fir is much less represented.
In the PF domain, performance on Douglas fir also varies strongly across models, as shown by the difference between its average and best scores in~\cref{fig:ind_f1_class_sp}.
We hypothesize that this variability is partly due to the very low number of samples of this class in this domain (\cref{fig:support_sp}). %Thus we add this exception of Douglas Fir in PF to the difficult class list in order to keep the cross-domain analysis results as clear as possible. % retiré car trop tot 

Second, F1 scores are lower in the TSA domain than in the other domains for most classes (69 on average, compared with more than 80 for the other domains), even though Spruce, which is difficult in this domain, is excluded from this average.
These lower scores may be related to the smaller number of samples in this domain (\cref{fig:support_sp}).
Within this domain, the ranking of the classes follows their number of samples, Spruce being both the least represented and the lowest-scoring class.

Apart from these two trends, Beech, Oak, Pine, and Larch obtain high in-domain F1 scores (above 74) in most domains.

% crops / For crop types mapping, w
\textbf{In-domain crop type mapping.} In~\cref{fig:ind_f1_class_tm}, the lowest scores are concentrated on three classes.
First, Winter Triticale is difficult to classify in all domains, one of the classes with the smallest support across domains.
It is mainly confused with Winter Wheat (confusion matrices in~\Cref{app:indomain}).
Both are winter cereals with similar phenological patterns.
This confusion has also been reported in other Sentinel-2 crop type studies~\cite{orynbaikyzy2020crop}.

Second, Winter Barley obtains lower F1 scores in the two French domains (FR1, FR2) than in DK and AUT.
Similarly to Winter Triticale, it is mainly confused with Winter Wheat in these two domains. 
%However it stays above the 50 mF1 points on average and 63 for the best encoder. 

Third, Spring Barley is well classified in DK, the only domain with a large number of samples for this class (\cref{fig:support_tm}), whereas its F1 score is significantly lower in the other domains, particularly in AUT and FR2.

Apart from these three classes, Corn, Meadow, Winter Rapeseed, and Winter Wheat obtain high in-domain F1 scores (above 87) in all domains.

% conclu
\textbf{Discussion.}
Overall, these results indicate that the benchmarked models achieve strong in-domain performance on both tasks.
The lower scores can be related to two main factors.
First, some classes are often confused with another class, such as Douglas fir with Pine, or Winter Triticale with Winter Wheat.
Second, other classes have few samples, such as Douglas fir in PF, Spruce in TSA, and Spring Barley outside DK.
Based on these observations, we identify the classes that are difficult in each domain with the following criterion: a class is difficult in a domain if its mean in-domain F1 score over all models is below 50, or if the F1 score of the best model is below 60.
For \specdataname, this criterion identifies Douglas fir in PolFI, and Spruce in TSA. We also add Douglas fir in PF, whose scores are unstable across models. 
For \cropdataname, it identifies Winter Triticale in all domains, and Spring Barley in AUT and FR2.
These class/domain pairs are hatched in~\cref{fig:ind_f1_class_sp,fig:ind_f1_class_tm}.
Since these classes are already poorly classified in-domain, they are unlikely to be well classified when they belong to the target domain.

Therefore, for a more reliable assessment of cross-domain generalization, as announced above, these classes are excluded from the cross-domain averages when their domain is the target.
Having identified these difficult classes, we now analyze the cross-domain performance of the models (\cref{subsec:bench_cross}).

%%%%%%%%%%%%%%%%%%%%% CROSS DOMAIN %%%%%%%%%%%%%%%%%%%
\vspace{0.4em}
\subsubsection{Cross-domain performance} \label{subsec:bench_cross}

We characterize the spatial generalization of \modelname and the baseline models on the two tasks and identify what drives their performance drops. For each task, we assess the cross-domain performance first at the domain level and then at the class level. Finally, we discuss the cross-domain performance of the benchmarked models.
We start with crop type mapping, whose cross-domain performance drops follow clearer patterns.

\begin{figure*}
    \centering
    \includegraphics[width=0.7\linewidth]{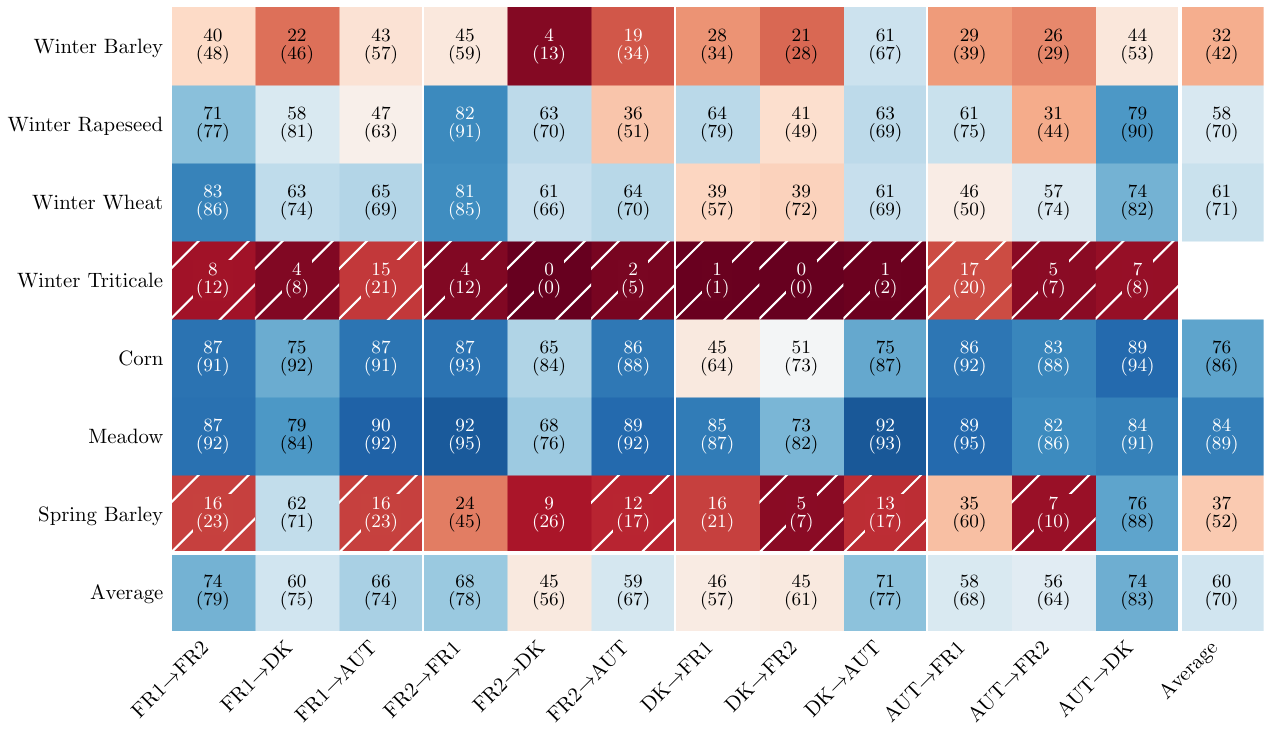}
     \caption{Cross-domain F1 per direction in \cropdataname, mean performance and max model performance in brackets. Color is centered on 50. Difficult classes of the target domain are hatched and excluded from the Average row and column. Presto is excluded.}
    \label{fig:cross_f1_class_tm}
\end{figure*}

\begin{figure}
    \centering
    \includegraphics[width=0.8\linewidth]{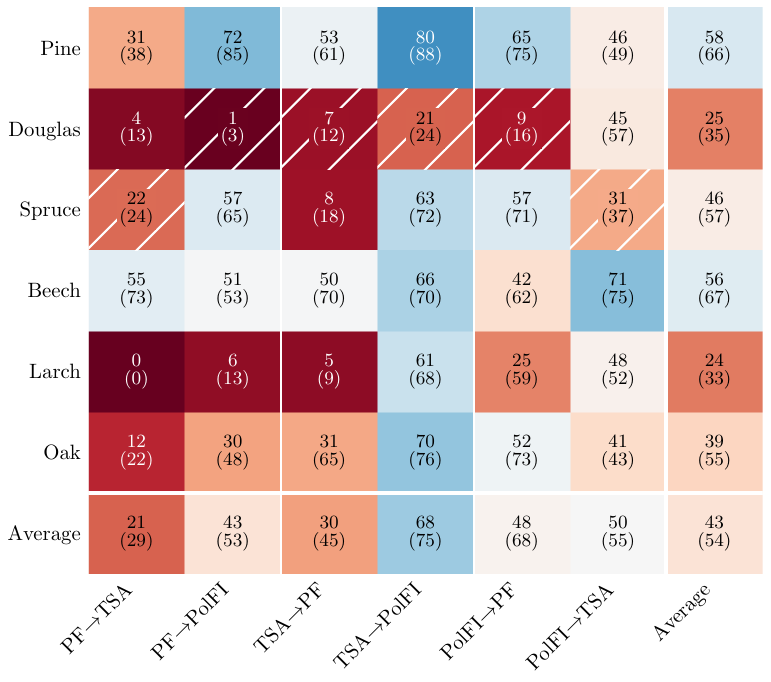}
     \caption{Cross-domain F1 per direction in \specdataname, mean performance and max model performance in brackets. Color is centered on 50. Difficult classes of the target domain are hatched and excluded from the Average row and column. Presto is excluded.}
    \label{fig:cross_f1_class_sp}
\end{figure}

\Cref{fig:cross_f1_class_sp,fig:cross_f1_class_tm} report the F1 score of each class for each direction (source\ra target), averaged over the same models as in~\cref{subsec:benchin_dom}.
The cross-domain score is computed on the test set of the target domain, and the drop is measured against the in-domain score of that same domain.
A class that is difficult in the target domain (\cref{subsec:benchin_dom}) is hatched in these figures.
It is excluded from the averages of the corresponding directions, since its in-domain score is not a reliable reference for computing the cross-domain drop\footnote{The averages in these figures therefore differ slightly from those in~\cref{tab:in_cross_sp,tab:in_cross_tm}, which include all classes. The same applies to the averages in the following figures and in the text.}.

\textbf{Cross-domain crop type mapping.} 
For crop type mapping, the average mF1 of the models drops from  85 in-domain to 60 cross-domain mF1 (average cells of~\cref{fig:ind_f1_class_tm,fig:cross_f1_class_tm}).
We relate this drop to two factors: the spatial distance between domains and the confusions already present in-domain.

\begin{figure}
    \centering
    \subfloat[\small \centering \specdataname \\ \label{fig:gap_dist_sp}]{%
        \includegraphics[width=0.49\linewidth]{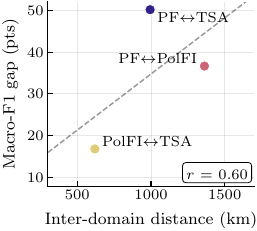}}
    \subfloat[\small \centering \cropdataname\\ \label{fig:gap_dist_tm}]{%
       \includegraphics[width=0.49\linewidth]{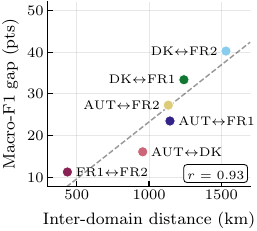}}
    \caption{Cross-domain mF1 drop versus the distance between domain centroids,
    averaged over the two directions of each pair and over the benchmarked models
    (Presto excluded). (a) \specdataname. (b) \cropdataname. $r$: Pearson correlation.}
    \label{fig:gap_dist}
\end{figure}

At the domain level, the cross-domain drop grows with the spatial distance between domains. \Cref{fig:gap_dist_tm} reports, for each domain pair, the drop averaged over its two directions as a function of the distance between the domain centroids: the correlation is strong across the six pairs.
We hypothesize that this relationship stems from differences in the timing of crop growth, which depends on the regional climate~\cite{nyborg2022generalized,nyborg2022timematch}.
At the class level, the classes that are best classified in-domain (Corn, Meadow, Winter Rapeseed, and Winter Wheat) remain the best classified cross-domain (\cref{fig:cross_f1_class_tm}).
Their lowest cross-domain scores are concentrated on the most distant pairs, which reflects the same distance effect.

\begin{figure}
    \centering
    \includegraphics[width=0.9\linewidth]{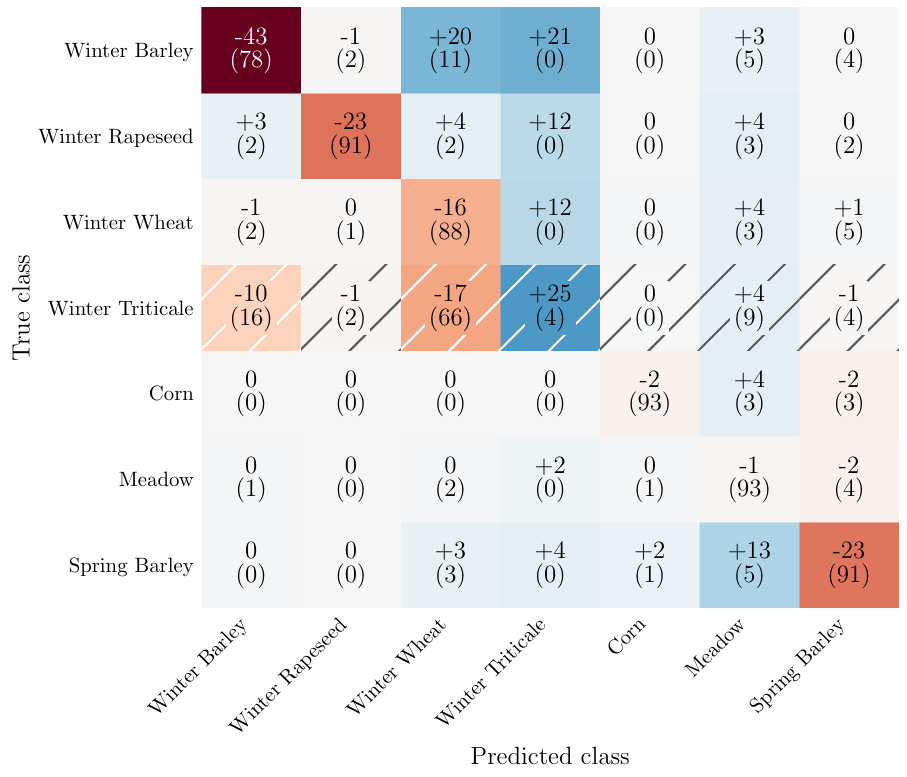}
  \caption{Difference between the AUT\ra DK and DK in-domain confusion matrices on \cropdataname, averaged over ALISE, AnySat, TESSERA-S2, and \modelname. Color is centered on 0. In-domain DK values are given in brackets.}
\label{fig:conf_aut_dk}
\end{figure}

In contrast, the classes with lower in-domain scores, such as Winter Barley and Spring Barley (\cref{subsec:benchin_dom}), perform even worse cross-domain. Their errors also extend to target domains where these classes were well classified in-domain.
For example, Winter Barley is well classified in-domain in DK and AUT, but obtains lower scores in the two French domains (\cref{fig:ind_f1_class_tm}). However, we observe a substantial drop in performance when transferring from AUT to DK. To identify the errors responsible for this drop, we compute the difference between the confusion matrix for the AUT \ra DK transfer and the in-domain confusion matrix for DK (\cref{fig:conf_aut_dk}).
In this transfer, Winter Barley loses more than half of its in-domain recall, and most of its misclassified pixels are predicted as Winter Wheat or Winter Triticale. Confusion with Winter Wheat already occurs in-domain in DK, but remains limited (11\% of the pixels). Cross-domain, it almost triples and becomes the main error for this class. In contrast, confusion with Winter Triticale is new, as it is absent in-domain in DK.

Overall, in addition to the distance effect, changing the source domain amplifies the confusions already present in-domain and extends them to domains where the class was initially well classified in-domain, with an intensity that grows with the distance between domains. 
It can also add new confusions, as observed on AUT\ra DK toward Winter Triticale, a class with more samples in the source domain than in the target domain.

% Corn bon exemple de pattern qui fonctionne en in-domain partout et qui drop seulement sur les domains eloignés 
%
% in domain c'etait: winter barley -> winter wheat sur FR1 et Fr2 (pas aut dk)   ====> only DK->AUT seems to work AUT->DK est la meilleur pair pourtant drop sur Winter Barley 
% en regardant confusion matrices  : goes to winter wheat comme le sautres 
% -> cross domain augmente les confusions deja presentes 
% indomain spring barley: low dans Fr2 et AUT mais bien dans FR1 et DK,
% DK donc fonctionne seulement en utilisant les domaines proches (AUT, FR1) mais aps FR2 le plus loin
% FR1 marche pas du tout: winter wheat agian
% dans les deux cas on dirait pas de rapport avec le support 
% en plus: 
% winter wheat -> falls ddown sur groupes phenologiques un peu ( pas de rapport avec support)
% winter rapeseed  -> sur les domains distants 
% corn meadow winter rapeseed and wheat work -> big support !

\textbf{Cross-domain tree species mapping.} 
Turning to the same analysis for tree species mapping, the average F1 of the benchmarked models drops from 78 mF1 in-domain to 43 cross-domain across the different domains (average cells of~\cref{fig:ind_f1_class_sp,fig:cross_f1_class_sp}), a larger drop (-35 points) than for crop types. 

At the domain level, unlike for crop types, the cross-domain drop does not clearly increase with the distance between domains (\cref{fig:gap_dist_sp}).
Two factors may explain this result.
First, \specdataname contains only three domains, which limits the number of pairs and the range of distances covered.
Second, the annotations differ between domains.
PF labels are strictly mono-specific annotations, whereas PolFI and TSA labels are multi-species stand-level inventory records assigned to their dominant species (\cref{subsec:downstream}).
The same class therefore does not strictly designate the same kind of object in PF and in the two other domains.
Consistent with this, the four directions involving PF have the lowest scores of the task (\cref{fig:cross_f1_class_sp}).
For instance, Larch is not a difficult class in any domain in the in-domain setting (\cref{fig:ind_f1_class_sp}), but its cross-domain F1 drops to 25 or less on the directions involving PF, while it remains higher between PolFI and TSA (61 and 48 F1).
This annotation mismatch may therefore hide the effect of distance on the pairs involving PF.

At the class level, we therefore focus on the PolFI\lra TSA pair, which is the spatially closest pair and the only one with the same type of annotations.
Even on this pair, every class that is not difficult in the target domain loses between 7 and 28 F1 points with respect to its in-domain score in the target (\cref{fig:ind_f1_class_sp,fig:cross_f1_class_sp}).
This drop also concerns classes that are not difficult in either domain in the in-domain setting, such as Pine and Beech.
This drop is therefore explained neither by in-domain difficulty nor by an annotation mismatch, and occurs even between the spatially closest domains.

\textbf{Discussion.} 
In this section, to isolate the cross-domain drops, we first set aside two factors that could otherwise explain them: the difficult classes of the target domain (\cref{subsec:benchin_dom}) and, for tree species, the annotation mismatch between PF and the other domains.

Once these factors are excluded, the benchmarked models still show three patterns of cross-domain drop: (i) for crop types, the drop grows with the distance between domains; (ii) changing the source domain in \cropdataname amplifies the confusions already observed in-domain and extends them to domains where the classes were well classified in-domain; (iii) for tree species, even on the closest pair with the same type of annotations, PolFI\lra TSA, each class that is not difficult in the target domain loses between 7 and 28 F1 points.

Lastly, the benchmarked models differ in their cross-domain behavior.
No single model is best on all target domains (\cref{tab:in_cross_sp,tab:in_cross_tm}).
On some directions, the best model clearly exceeds the average of the models, as for Winter Rapeseed on FR1\ra DK (81 versus 58 F1,~\cref{fig:cross_f1_class_tm}).
However, these advantages are limited to specific directions and classes.
Some baseline models may reduce the drop on some directions without improving spatial generalization entirely.
The in-depth analysis performed in this section also indicates that averaged cross-domain scores give an incomplete view of spatial generalization, as they hide large differences between directions and between classes of the same direction.
%The next section examines whether auxiliary variables can reduce these cross-domain patterns.

%%%%%----- AUXILIARIES CONTRIBUTION (VANILLA VS AUX)----%%%

\subsection{Contribution of auxiliary variables} \label{subsec:aux_res}

In this section, we examine whether the integration of auxiliary variables reduces the three cross-domain drop patterns identified in~\cref{subsec:bench_cross}.
We compare the variants presented in~\cref{subsec:aux-integration} with \modelname.
Each variant integrates one of the era5, soil, and topo sources, either as additional input tokens (\modelnamein) or as additional pre-training targets (\modelnameout).
As presented in~\cref{subsec:aux}, combo denotes the \modelnameout variant that combines the three auxiliary sources.
We measure the effect of each variant by \dvan, the difference between the F1 of the variant and that of \modelname on the same transfer direction.
\Cref{tab:in_cross_sp,tab:in_cross_tm} report the in-domain and cross-domain performance of the variants \modelnameout and \modelnamein with each auxiliary source and with their combination for \modelnameout.
We first identify the most effective integration strategy for spatial generalization (\cref{subsec:aux-integration_res}).
We then examine, for each task, how the retained strategy acts on each identified pattern (\cref{subsubsec:aux_patterns}).

\vspace{0.4em}
\subsubsection{Integration strategy}\label{subsec:aux-integration_res}

We first compare the two integration strategies introduced in~\cref{sec:method}: auxiliary variables used as additional inputs (\modelnamein) or as additional targets (\modelnameout).
Overall, our results corroborate the findings of In-N-Out~\cite{xie2021innout}: auxiliary variables used as inputs tend to improve in-domain performance, whereas auxiliary variables used as targets tend to improve cross-domain performance.

In-domain, \modelnamein variants reach the best score among frozen models in six of the seven target domains (\cref{tab:in_cross_sp,tab:in_cross_tm}).
These in-domain gains can be large.
For instance, \inera reaches the best average in-domain mF1 of all frozen models on both tasks, with +8.3 mF1 over \modelname on \specdataname.
Used as targets, the auxiliary variables also tend to improve the in-domain scores on \specdataname, but not on \cropdataname.
This suggests that the auxiliary targets are more useful for tree species mapping than for crop type mapping, possibly because the former task is harder in-domain and has fewer labeled samples (\cref{subsec:benchin_dom}). %whereas the extra information added in \modelnamein may only add noise for the inference on crop types.

Cross-domain, the trend is reversed: among the \modelname variants, \modelnameout variants reach the best score in six of the seven target domains across the two tasks, while \inera, the best in-domain \modelname variant, is the weakest variant on both tasks.
Following In-N-Out, we hypothesize that a linear layer trained on frozen representations that contain auxiliary values may exploit relations between these values and the classes, and that these relations may not hold in other domains.
On crop types, the cross-domain gain of \modelnameout comes mainly from the directions toward AUT, which we analyze in~\cref{subsubsec:aux_patterns}.
Additionally, \modelnameout variants reach the best average cross-domain mF1 of all frozen models on both tasks, with fewer parameters than AnySat and TESSERA-S2, and without requiring auxiliary data at inference.
Because we focus our analysis on spatial domain generalization, we retain the target integration for the rest of the analysis.

\begin{figure*}[]
    \centering
    \includegraphics[width=0.6\linewidth]{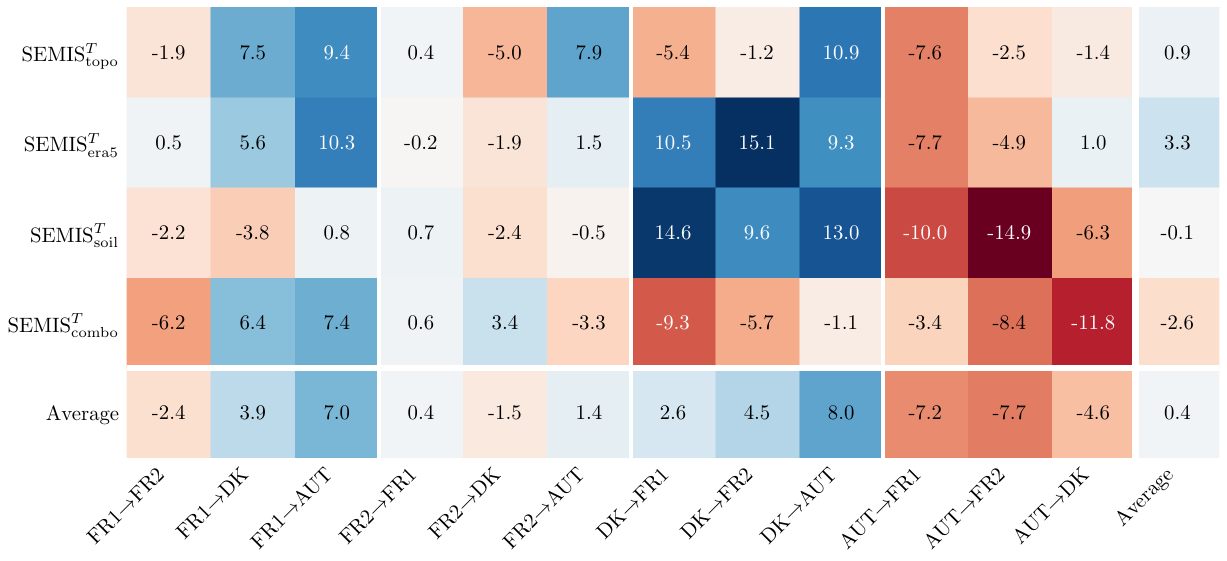}
        \caption{\dvan of each \modelnameout variant (rows) for each direction (source\ra target) of \cropdataname (columns). \dvan is the F1 difference between the variant and \modelname, averaged over the classes that are not difficult in the target domain. The Average row and column give the mean over variants and over directions. Blue cells indicate that the variant outperforms \modelname and red cells that it underperforms, with a color scale centered on 0. Because difficult classes are excluded, these averages differ from the mF1 differences of~\cref{tab:in_cross_tm}.}
    \label{fig:cross_delta_by_dir_tm}
\end{figure*}

\begin{figure}[]
    \centering
    \includegraphics[width=0.9\linewidth]{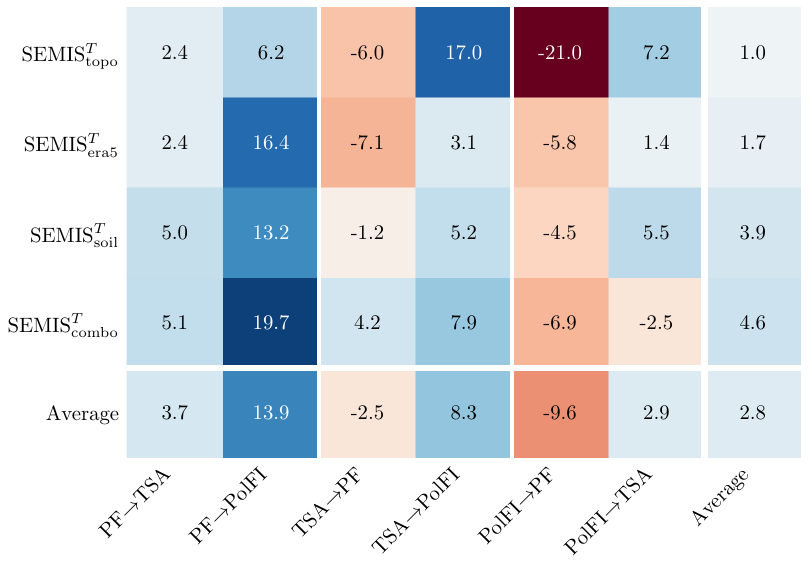}
    \caption{\dvan of each \modelnameout variant (rows) for each direction (source\ra target) of \specdataname (columns). \dvan is the F1 difference between the variant and \modelname, averaged over the classes that are not difficult in the target domain. The Average row and column give the mean over variants and over directions. Blue cells indicate that the variant outperforms \modelname and red cells that it underperforms, with a color scale centered on 0. Because difficult classes are excluded, these averages differ from the mF1 differences of~\cref{tab:in_cross_sp}.}
    \label{fig:cross_delta_by_dir_sp}
\end{figure}

\vspace{0.4em}
\subsubsection{Identified patterns}\label{subsubsec:aux_patterns}
We now examine how \modelnameout acts on the three patterns identified in~\cref{subsec:bench_cross}. We first focus on the failures identified on \cropdataname, then we turn to \specdataname.

\textbf{Crop type mapping.} We focus on pattern (i), the increase of the cross-domain drop with the distance between crop domains.
\Cref{fig:cross_delta_by_dir_tm} reports \dvan for each \modelnameout variant and each direction of \cropdataname, averaged over the classes that are not difficult in the target domain.
Our results show that several of the largest gains can be related to specific environmental differences between the two domains that the auxiliary variables describe. However, no auxiliary source improves all directions.
For instance, topography (\outtopo) gives the most regular gain on the three directions toward AUT (average \dvan of +9.4). This can be related to AUT being the only alpine domain of \cropdataname.
Climate and soil (\outera, \outsoil) improve all three directions from DK, whereas topography improves only DK\ra AUT. Moreover, without auxiliary variables, DK\ra FR1 and DK\ra FR2 are among the hardest transfer pairs~(\cref{fig:cross_f1_class_tm}). However, \outera and \outsoil improve these two directions, with \dvan ranging from +9.6 to +15.1. This is consistent with DK differing from the French domains (FR1 and FR2) in climate and soil.%, while none of these three domains is mountainous.

These gains are therefore obtained where pattern (i) is the most severe.
However, the directions with AUT as the source domain are degraded by most variants.
This asymmetry shows that an auxiliary source that helps in one direction of a domain pair does not necessarily help in the other.

\input{sections/tables/recovered_aux}

Next, we focus on pattern (ii), the extension of in-domain confusions to domains where the class was well classified in-domain.
Since it concerns specific classes, we examine it at the class level.
A class/direction pair denotes a class evaluated on one transfer direction.
We consider the 23 pairs of \cropdataname whose F1, averaged over the models, falls below 50 and whose class is not difficult in the target domain (\cref{fig:cross_f1_class_tm}).
\Cref{tab:recovered} reports, for each class, the number of pairs on which the best \modelnameout variant outperforms \modelname and the best baseline model, together with the pair on which it most exceeds \modelname.
When the best auxiliary source is chosen for each pair, \modelnameout improves 19 of the 23 identified pairs over \modelname, with an auxiliary source that changes across pairs.
It also outperforms the best baseline model on 14 of these 23 pairs.
The clearest case is Winter Barley on AUT\ra DK, the example used to illustrate pattern (ii): with \outera, its F1 rises from 41.2 to 70.1, above the best baseline model.
Spring Barley, the other class cited for pattern (ii) in~\cref{subsec:bench_cross}, also benefits from the auxiliary variables.
On FR2\ra DK, where it drops sharply despite being well classified in DK in-domain, its F1 rises from 3.9 to 58.6 with \outcombo.
Nonetheless, four pairs out of 23 are degraded compared to \modelname without auxiliary variable integration (three of them with AUT as the source domain), consistent with the degradation observed at the direction level~(\cref{fig:cross_delta_by_dir_tm}).

\textbf{Tree species mapping.} 
We now focus on pattern (iii), that is, the cross-domain drop between TSA and PolFI, the spatially closest tree species domains with the same type of annotations.
At the direction level, every \modelnameout variant improves TSA\lra PolFI on average over the two directions (\cref{fig:cross_delta_by_dir_sp}).
Topography is the most effective auxiliary source on this pair (average \dvan of +12.1), with a \dvan of +17.0 on TSA\ra PolFI.
This may be related to the role of elevation in forest composition in Central Europe~\cite{leuschner2017ecology}, which the topographic source encodes directly.
At the class level, pattern (iii) concerns all the classes of the two directions rather than specific classes.
We therefore consider the 10 class/direction pairs of TSA\lra PolFI (Douglas fir and Spruce are excluded as difficult classes in PolFI and TSA, respectively).
The best \modelnameout variant improves 9 of them over \modelname, mostly with topography, and outperforms the best baseline model on all 10 (\cref{tab:recovered}).

Finally, beyond this restricted pattern in \specdataname, the most consistent gains on \specdataname are obtained with PF as the source domain: all \modelnameout variants improve both directions from PF (\cref{fig:cross_delta_by_dir_sp}).
As PF follows a different type of annotations, these gains cannot be attributed to geographic differences alone.
They nonetheless suggest that auxiliary variables used as targets can also help transfer between domains with different types of annotations. That is a frequent situation in practice since reference data from different countries follow different annotation protocols. % although not towards PF.
However, this gain only holds in one direction, from the mono-specific labels of PF to the mixed-species labels of PolFI and TSA, and not the reverse.

\textbf{Discussion.}
Taken together, these results support a recommendation on the integration strategy, but not a single choice of auxiliary source.
On the one hand, auxiliary variables used as inputs give the best in-domain scores, whereas auxiliary variables used as pre-training targets give the best cross-domain scores among the \modelname variants, consistent with In-N-Out~\cite{xie2021innout}.
When the objective is transfer to unseen regions with frozen encoders, our results therefore favor auxiliary variables used as targets.
This choice is also simpler at the operational level, since targets leave inference unchanged, whereas inputs require the auxiliary source to be available over the whole mapped area.
On the other hand, the choice of source is less clear, since no auxiliary source improves all directions.
Each of the three patterns is mitigated by at least one auxiliary source, and several of the largest gains can be related to specific environmental differences described by the auxiliary variables, such as topography toward AUT and on TSA\lra PolFI directions, and climate and soil from DK.
The results are not systematically symmetric: with auxiliary variables as targets, the gains depend on the direction, and some directions are degraded, in particular with AUT as the source domain.
%As a result, selecting the best domain source for a given direction would require labels in the target domain.

%% file: sections/tables/in_cross_sp.tex
\begin{table*}[t]
\caption{In-domain and cross-domain macro F1 (mF1) of each model on the test set of each \specdataname domain. For each target domain, the cross-domain score is averaged over the source domains. The Avg. columns are averaged over the target domains and include all classes. The best and second-best frozen models are highlighted in \colorbox{bestgreen}{\textbf{green}} (best also in bold) and \colorbox{secondgreen}{light green}. \modelnameFS is \underline{underlined} when it outperforms the best frozen model. The numbers of frozen (\textcolor{cyan}{\faSnowflake}) and trainable (\textcolor{orange}{\faFire}) parameters are given for each model.}
\centering
\label{tab:in_cross_sp}
\begin{tabular}{l@{\hspace{3pt}}|@{\hspace{3pt}}r@{\hspace{5pt}}r@{\hspace{3pt}}|cc|cc|cc|cc}
\toprule
 & \multicolumn{2}{c}{parameters} & \multicolumn{2}{c}{PolFI} & \multicolumn{2}{c}{TSA} & \multicolumn{2}{c}{PF} & \multicolumn{2}{c}{Avg.} \\
 &  \textcolor{cyan}{\faSnowflake} &\textcolor{orange}{\faFire} & In & Cross & In & Cross & In & Cross & In & Cross \\
\midrule
\modelnameFS &0 &11.3M& 80.39 & 32.20 & 72.75 & 32.63 & 81.69 & 31.01 & 78.28 & 31.95 \\
\midrule
Presto &$404$K&768 &  68.39 & 16.51 & 60.14 & 31.88 & 28.13 & 17.21 & 52.22 & 21.87 \\
ALISE &$1.1$M&3.8K & 69.45 & 42.48 & 62.43 & 27.09 & 66.33 & 21.37 & 66.07 & 30.31 \\
TESSERA-S2 &$87.8$M&3.1K & 73.37 & 52.86 & 64.88 & 36.19 & 83.98 & 35.44 & 74.08 & 41.50 \\
AnySat &$125$M&4.6K & 77.57 &  53.39 & 65.43 & 36.09 & 80.44 & 34.29 & 74.48 & 41.26 \\
\modelname &$11.1$M&2.3K &77.12 & 43.80 & \cellcolor{secondgreen}71.67 & 36.13 & 65.89 & \cellcolor{secondgreen}44.20 & 71.56 & 41.38 \\
\textit{Avg.} & & & \avgv{73.18} & \avgv{41.81} & \avgv{64.91} & \avgv{33.48} & \avgv{64.95} & \avgv{30.50} & \avgv{67.68} & \avgv{35.26} \\
\midrule
\modelnamein\, topo& $11.1$M & 2.3K  & 80.53 & 47.99 & 71.25 & 38.25 & 80.13 & \cellcolor{bestgreen} \textbf{45.54} & 77.30 & 43.92 \\
\hphantom{\modelnamein}\, era5&& & 76.14 & 42.71 & \cellcolor{bestgreen}\textbf{73.09} & 35.52 & \cellcolor{bestgreen}\textbf{90.48} & 33.46 & \cellcolor{bestgreen}\textbf{79.90} & 37.23 \\
\hphantom{\modelnamein}\, soil & & & 77.82 & 48.26 & 69.68 & 38.28 & 88.16 & 27.71 & 78.55 & 38.08 \\
\textit{Avg.} & & &\avgv{78.16} &  \avgv{46.32} &  \avgv{71.34} &  \avgv{37.35} &  \avgv{86.26} &  \avgv{35.57} &  \avgv{78.59} &  \avgv{39.75} \\
\greymidrule
\modelnameout\, topo & $11.1$M & 2.3K & \cellcolor{bestgreen}\textbf{81.63} & \cellcolor{secondgreen}53.79 & 71.30 & \cellcolor{secondgreen} 40.70 & 61.31 & 32.58 & 71.42 & 42.36 \\
\hphantom{\modelnameout}\, era5&&& 75.75 & 51.86 & 70.55 & 39.62 & 78.81 & 37.84 & 75.04 & 43.11 \\
\hphantom{\modelnameout}\, soil &&& 80.18 & 51.99 & 70.07 & \cellcolor{bestgreen}\textbf{42.31} & \cellcolor{secondgreen} 88.90 & 40.93 & \cellcolor{secondgreen} 79.72 & \cellcolor{secondgreen} 45.08 \\
\hphantom{\modelnameout}\, combo &&& \cellcolor{secondgreen} 81.02 & \cellcolor{bestgreen}\textbf{54.28} & 69.46 & 39.85 & 84.93 & 42.41 & 78.47 & \cellcolor{bestgreen}\textbf{45.51} \\
\textit{Avg.} & & &\avgv{79.65} & \avgv{52.98} & \avgv{70.35} & \avgv{40.62} & \avgv{78.49} & \avgv{38.44} & \avgv{76.16} & \avgv{44.01} \\
\bottomrule

\end{tabular}
\end{table*}

%% file: sections/tables/in_cross_tm.tex
\begin{table*}[t]
\caption{In-domain and cross-domain macro F1 (mF1) of each model on the test set of each \cropdataname domain. For each target domain, the cross-domain score is averaged over the source domains. The Avg. columns are averaged over the target domains and include all classes. The best and second-best frozen models are highlighted in \colorbox{bestgreen}{\textbf{green}} (best also in bold) and \colorbox{secondgreen}{light green}. \modelnameFS is \underline{underlined} when it outperforms the best frozen model. The numbers of frozen (\textcolor{cyan}{\faSnowflake}) and trainable (\textcolor{orange}{\faFire}) parameters are given for each model.}
\label{tab:in_cross_tm}
\centering
\begin{tabular}{l@{\hspace{3pt}}|@{\hspace{3pt}}r@{\hspace{5pt}}r@{\hspace{3pt}}|cc|cc|cc|cc|cc}
\toprule
 & \multicolumn{2}{c}{parameters} & \multicolumn{2}{c}{AUT} & \multicolumn{2}{c}{DK} & \multicolumn{2}{c}{FR1} & \multicolumn{2}{c}{FR2}  & \multicolumn{2}{c}{Avg.} \\
 & \textcolor{cyan}{\faSnowflake} &\textcolor{orange}{\faFire} & In & Cross & In & Cross & In & Cross & In & Cross  & In & Cross \\
\midrule
\modelnameFS &0 &11.4M& \underline{88.48} & \underline{58.96} & \underline{92.00} & \underline{56.76} & \underline{92.22} & \underline{60.56} & \underline{83.18} & \underline{48.34} & \underline{88.97} & \underline{56.16} \\
\midrule
Presto &$404$K&896 & 53.58 & 15.57 & 61.55 & 7.30 & 48.53 & 18.43 & 38.12 & 15.60 & 50.44 & 14.23 \\
ALISE &$1.1$M&4.5K & 74.48 & 46.33 & 75.02 & 44.31 & 73.93 & 47.95 & 58.66 & 41.21 & 70.52 & 44.95 \\
TESSERA-S2 &$87.8$M&3.6K & 75.21 & \cellcolor{secondgreen}53.31 & 75.46 & \cellcolor{secondgreen}54.11 & 71.40 & 48.89 & 60.24 & \cellcolor{bestgreen}\textbf{48.03} & 70.58 & \cellcolor{secondgreen}51.09 \\
AnySat &$125$M&5.4K & 78.08 & 52.23 & 79.18 & \cellcolor{bestgreen}\textbf{55.74} & 75.20 & 49.48 & 62.96 & 41.86 & 73.85 & 49.83 \\

\modelname &$11.1$M&2.7K & \cellcolor{secondgreen}81.35 & 45.77 & 80.70 & 52.57 & 81.16 & 54.59 & 69.53 & 42.82 & 78.18 & 48.94 \\
\textit{Avg.} & & &  \avgv{72.54} & \avgv{42.64} & \avgv{74.38} & \avgv{42.81} & \avgv{70.04} & \avgv{43.87} & \avgv{57.90} & \avgv{37.90} & \avgv{68.72} & \avgv{41.81} \\
\midrule
\modelnamein\, topo &$11.1$M&2.7K & 81.06 & 50.18 & 80.99 & 51.87 & \cellcolor{secondgreen}81.36 & 51.27 & \cellcolor{secondgreen}69.77 & 41.12 & \cellcolor{secondgreen}78.30 & 48.61 \\
\hphantom{\modelnamein}\, era5 &&& \cellcolor{bestgreen}\textbf{83.01} & 43.27 & \cellcolor{bestgreen}\textbf{83.47} & 46.83 & \cellcolor{bestgreen}\textbf{83.70} & 51.59 & \cellcolor{bestgreen}\textbf{71.31} & 42.88 & \cellcolor{bestgreen}\textbf{80.37} & 46.14 \\
\hphantom{\modelnamein}\,  soil &&& 81.04 & 48.70 & \cellcolor{secondgreen}81.52 & 53.47 & 81.18 & 51.56 & 67.55 & 37.84 & 77.82 & 47.90 \\
\textit{Avg.} & & &  \avgv{81.70} & \avgv{47.38} & \avgv{81.99} & \avgv{50.72} & \avgv{82.08} & \avgv{51.47} & \avgv{69.54} & \avgv{40.61} & \avgv{78.83} & \avgv{47.55} \\
\greymidrule
\modelnameout\, topo &$11.1$M&2.7K& 80.27 & \cellcolor{bestgreen}\textbf{55.08} & 81.21 & 53.15 & 80.73 & 50.73 & 67.66 & 41.76 & 77.47 & 50.18 \\
\hphantom{\modelnameout}\,  era5 &&& 79.52 & 51.34 & 79.97 & \cellcolor{secondgreen}54.11 & 80.42 & \cellcolor{secondgreen}54.72 & 68.51 & \cellcolor{secondgreen}45.68 & 77.11 & \cellcolor{bestgreen}\textbf{51.46} \\
\hphantom{\modelnameout}\,  soil &&& 79.95 & 49.03 & 80.79 & 49.10 & 80.77 & \cellcolor{bestgreen}\textbf{56.21} & 67.86 & 40.77 & 77.34 & 48.77 \\
\hphantom{\modelnameout}\, combo &&& 79.83 & 47.81 & 80.35 & 52.26 & 80.28 & 50.32 & 67.64 & 37.97 & 77.02 & 47.09 \\
\textit{Avg.} & & &  \avgv{79.89} & \avgv{50.81} & \avgv{80.58} & \avgv{52.15} & \avgv{80.55} & \avgv{53.00} & \avgv{67.92} & \avgv{41.54} & \avgv{77.23} & \avgv{49.38} \\
\bottomrule
\end{tabular}

\end{table*}

%% file: sections/tables/recovered_aux.tex
\begin{table}
\centering
\footnotesize
\setlength{\tabcolsep}{2pt}
\caption{Effect of \modelnameout on the pattern (ii) pairs of \cropdataname (average F1 below 50) and on the TSA$\leftrightarrow$PolFI pairs of \specdataname (pattern (iii)), excluding classes difficult in the target. For each pair, the best \modelnameout variant is kept. $n$: number of pairs. Best baseline: best of ALISE, AnySat, and TESSERA-S2. Best case: largest gain over \modelname. Per-pair results in~\cref{app:recovered_tm,app:recovered_sp}.}
\label{tab:recovered}
\begin{tabular}{lcccl}
\toprule
 & & \multicolumn{2}{c}{Improved over} & Best case \\
\cmidrule(lr){3-4}
Class & $n$ & \modelname & Best base. & \modelname\,\ra\,\modelnameout \\
\midrule
\multicolumn{5}{l}{\textit{\cropdataname}} \\
W. Barley   & 11 & 10 & 6  & AUT\ra DK (era5): 41.2\,\ra\,70.1 \\
W. Wheat    & 3  & 3  & 1  & DK\ra FR1 (soil): 5.2\,\ra\,41.4 \\
W. Rapeseed & 4  & 2  & 2  & FR1\ra AUT (combo): 36.2\,\ra\,73.8 \\
S. Barley   & 4  & 3  & 4  & FR2\ra DK (combo): 3.9\,\ra\,58.6 \\
Corn        & 1  & 1  & 1  & DK\ra FR1 (era5): 21.6\,\ra\,79.8 \\
\cmidrule(lr){1-5}
All classes & 23 & 19 & 14 &  \\
\midrule
\multicolumn{5}{l}{\textit{\specdataname} (TSA\lra PolFI only)} \\
All classes & 10 & 9  & 10 & TSA\ra PolFI (topo): 64.7\,\ra\,81.7  \\
\bottomrule
\end{tabular}
\end{table}

%% file: sections/7_conclusion.tex
\label{sec:conclu}
%% context 
%probleme, produire des cartes,generalization 
    %scope FM SITS utilisées gelées -> robustness spatial doit venir du pre-training
This work addresses the spatial generalization capacity of SITS FMs required for large-scale vegetation mapping. In such a scenario, where labels are scarce, shallow models trained on SITS representations from one spatial domain should ideally be applicable to another domain.

In this paper, we evaluate the spatial generalization performance of pre-trained and frozen SITS FMs that provide fixed-size, easy-to-use representations. In addition, as a growing number of SITS FM studies integrate auxiliary variables either as inputs or as pre-training targets, we study their influence on spatial generalization. To do so, we propose \modelname, a reference SITS encoder pre-trained as a masked autoencoder, with auxiliary variables used either as inputs (\modelnamein) or as targets (\modelnameout).
%protocol
    %spatial generalization
%data
    %SPAN : TSA, PolFI, PF, shift spatial et inter-annuel
    %RapidCropsSG : tuiles TimeMatch, shift purement spatial
%models
    %baselines frozen : ALISE, AnySat, TESSERA-S2 (Presto exclu des moyennes)
    %SEMIS : backbone de controle, suivant des sota pour l'archi , fixed size embedding
    %auxiliaires : climat, sol, topographie
    %intégration input ou target
Furthermore, we propose a spatial generalization framework in which a single linear classifier is trained on top of the SITS representations in one spatial domain and tested in another domain, without using any data from the target domain.
This evaluation is performed on two tasks using two benchmark datasets, \specdataname, specifically designed for the evaluation of spatial generalization of tree species mapping, and \cropdataname, an extension of the TimeMatch dataset. 
We evaluate the spatial generalization of four baseline encoders using the same Sentinel-2 (S2) SITS inputs: Presto, ALISE, AnySat, and TESSERA-S2~\cite{tseng2023lightweight,dumeur2025paving,astruc2025anysat,feng2025tessera}, as well as our reference SITS encoder \modelname and its two auxiliary-variable variants, \modelnamein and \modelnameout.

%% Where does this performance drop occur under spatial shift, and what does it depend on?
%%%% faire crops vs trees 
%selectionner
%citer chiffre de baisse moyenne de mF1 entre in-domain et cross-domain sur les deux tasks 
%distance
    %drop augmente avec distance 
    %surtout visible sur les cultures
%confusions, pas trop precis
    %confusion in-domain -> sur nouveaux domains ou ca n'arrivait pas en in-domain
    %spread nouvelles confusions dans classes avec bons support source -> vrai shift
%ecart résiduel sur SPAN
    %hors PF, TSA/PolFI reste nettement sous la référence in-domain
%? parler de SSL vs FT ? SSL >  FS cross-domain sur les arbres, pas sur crops?
The mF1 of every baseline and of \modelname dropped by 30 to 36 points from in-domain to cross-domain for tree species, and by 19 to 36 points for crop types.
For crop types, this drop grew with the spatial distance between domains.
In addition, confusions already present in-domain were amplified and extended to other domains where the class was well classified in-domain.
For tree species, the transfers involving PureForest were the weakest, possibly because its annotations describe only mono-specific forests, whereas the two other datasets assign each stand to its dominant species. 
Other transfers in the tree species benchmark still exhibit large per-class F1 score drops, which differ strongly from one class to another,  ranging from 7 to 28 F1 points.
%Even between these two remaining regions, which are spatially closest region, each class lost between 7 and 28 F1 points.
These differences motivated our analysis of spatial generalization at finer granularity such as per direction and per class, rather than with a single averaged score.

Finally, our self-supervised pre-trained \modelname, used as a frozen encoder, outperformed the fully supervised \modelnameFS in cross-domain evaluation on tree species. This result suggests that pre-training benefits spatial generalization and favors the development of FMs for large-scale tree species map production, where labels are particularly scarce.

%% Does auxiliary integration reduce this gap, and does it matter whether the auxiliaries are integrated as inputs or as targets?
%integration: reprendre NEURIPS 
    %input
        %meilleur in-domain du benchmark
        %pire variante SEMIS en cross-domain
    %target
        %améliore le transfert dans la majorité des configurations, dont toutes celles de SPAN
        %meilleur modèle gelé en cross-domain sur les deux tâches
        %moins de paramètres qu'AnySat et TESSERA-S2
        %aucun AVs à l'inférence
%source: dire effets in domain gros boost (era5), mais pas de grande tendance reliées aux sources 
%gains precis sur classes echecs dans benchmarks
    %la plupart des paires classe/direction en échec sont améliorées par rapport à vanilla, sur les deux tâches
%Regarding the second question, 
Regarding the integration strategy of auxiliary variables, \modelnamein (auxiliary variables as input) obtained the best in-domain scores, whereas \modelnameout (auxiliary variables as target) obtained better cross-domain scores than both \modelname and \modelnamein. % consistent with In-N-Out~\cite{xie2021innout}.
Among all frozen FMs, baselines included, \modelnameout also reached the highest average cross-domain mF1 on both tasks, although by a small margin on crop types. Moreover, unlike \modelnamein, it does not require auxiliary data at inference.
When the best auxiliary source was selected for each class/direction pair, \modelnameout reduced most of the class-level drops described above, on both tasks.
Some of the largest gains might be related to environmental differences between domains. For instance, pre-training with topography as auxiliary target seems to improve transfers toward the alpine Austrian region.
However, no auxiliary variable source improved all transfer directions, and some transfers were degraded. 
In particular, transfers from the Austrian region were degraded.
This exception does not appear in the scores averaged over all directions, in which auxiliary targets give the best results, and is only revealed by the per-direction evaluation.
To conclude, our results support using auxiliary variables as pre-training targets to improve the spatial generalization of SITS FMs, while the choice of the auxiliary source remains open.

%% limites/perspectives
%closed-set/harmonization
    %moins de labels
    %pas réaliste sur l'utilisation end-user
    %perte de granularité (car diferentes entre les datasets)
    %closed-set/harmonization
    %open-set
    %granularité plus fine
    %datasets utilisables tels quels ! 
%auxiliaires il y a plein d'autres et pas agrégés dans le temps
	%auxiliaires agrégés
    %auxiliaires à résolution temporelle
    %autres auxiliaires 
%% limites/perspectives
Our study opens up several possibilities for future research, which arise from its two main limitations. 
First, our evaluation relies on a closed set of classes shared by all spatial domains. 
Although this harmonization was necessary to compare models across domains, it discarded the non-shared classes, thereby reducing both the number of labels and the class granularity.
In practice, however, end users would rather keep their own nomenclature, so that the target domain may contain classes that are absent from the source domain and vice versa.
Evaluating spatial generalization in an open-set setting~\cite{panareda2020}, where the model must classify the shared classes and identify target samples from unseen classes as unknown, is therefore a natural next step toward fully addressing spatial generalization.
Second, only three auxiliary sources were tested, and all were encoded as temporally static variables (\cref{subsec:aux}).
Temporally resolved auxiliary variables, such as daily climate data aligned with the SITS acquisitions, as well as additional auxiliary sources, could better characterize the differences between domains.
%Second, we tested only three auxiliary sources, and the climate variables were averaged temporally over each SITS period, which removed their temporal information.
%Temporally resolved auxiliary variables, such as daily climate information aligned with the SITS acquisitions, and other auxiliary sources could describe further the differences between regions.

%More generally, the proposed datasets, protocol and analysis provide a basis to assess whether future SITS FMs can be applied as is to unseen regions.

%% file: sections/X_appendix.tex
\subsection{\modelname attention layers}

Following L-TAE~\cite{garnot2020lightweight}, the values are not projected in the cross-attention of the temporal summarization (\cref{subsec:model}).
In each decoder block (\cref{subsec:pre-training}), the queries are layer-normalized before the cross-attention, which is wrapped in a residual connection. The cross-attention is followed by a GEGLU feed-forward network with its own layer normalization and residual connection, except in the last block, where the feed-forward network is omitted. As in the temporal summarization, the values are not projected. 
\subsubsection{Local spatial attention}
\begin{figure}
    \centering
    \includegraphics[width=0.7\linewidth]{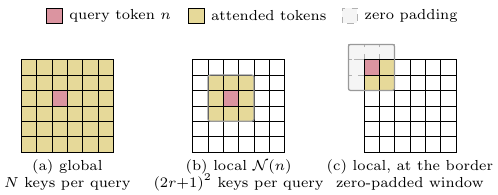}
   \caption{Spatial attention at a fixed acquisition $t$. (a) Global attention of the original TimeSformer. (b) Local attention used in \modelname, restricted to the $3\times3$ neighborhood $\mathcal{N}(n)$. (c) At the border, the window is completed by zero padding.}
    \label{fig:local_att}
\end{figure}

\label[appendix]{app:local-attention}
In the original TimeSformer, the spatial attention of \cref{eq:space_at} is global: at a fixed acquisition $t$, each token attends to the $N$ tokens of the same image (\cref{fig:local_att} (a)). As stated in \cref{sec:method}, our objects of interest are static (trees and crops) within the input time series (1 to 2 years) and mostly characterized by their local context. Therefore, we restrict the spatial attention layer to the neighborhood $\mathcal{N}(n)$, the $(2r+1)\times(2r+1)$ window of patches centered on $n$, with $r=1$ (\cref{fig:local_att} (b)). 

Technically, keys and values are gathered with a sliding window, and each query attends only to the $(2r+1)^2=9$ tokens of its window. With $64\times64$ inputs and $P=4$, $N=256$: each query compares 9 keys instead of 256. At the image borders, the window is completed by zero padding (\cref{fig:local_att} (c)). After the $L_{enc}=8$ blocks, the receptive field spans $2rL_{enc}+1=17$ patches, i.e., $68$ pixels, which covers the whole $16\times16$ patch grid. 

\subsection{Label harmonization}
\label[appendix]{app:harmonization}
\subsubsection{\specdataname harmonization}

Each of the three source datasets comes with its own nomenclature, defined by its authors at either the genus or the species level (16 classes for PolFI, 20 for TSA, and 13 for PF). We map every published class to its genus, and keep the genera present in the three datasets. \cref{tab:span-harmonization} details the mapping from the published labels to the \specdataname classes.

Six classes are retained. Pine, Spruce, Larch, and Beech group all the
species of the corresponding genus, while Oak groups only its deciduous species. Genera present in only one or two datasets are discarded. Fir is present in the three datasets but is not retained, as detailed in \cref{tab:span-harmonization}. Pixels of all discarded classes are set to the no-data value and ignored during probing and evaluation.

\input{sections/tables/spec_mapping}

\subsubsection{\cropdataname harmonization}
\cropdataname labels are derived from the HCAT3 classes of RapidCrops~\cite{rapidcrops}.
A class is retained if it is present in the four domains and covers at least 0.4\% of the labeled pixels of each domain. Seven classes meet this criterion: Corn, Meadow, Spring Barley, Winter Barley, Winter Rapeseed, Winter Triticale, and Winter Wheat. Pixels of all other classes are set to the no-data value and are ignored during probing and evaluation.

\begin{figure}
    \centering
    \includegraphics[width=\linewidth]{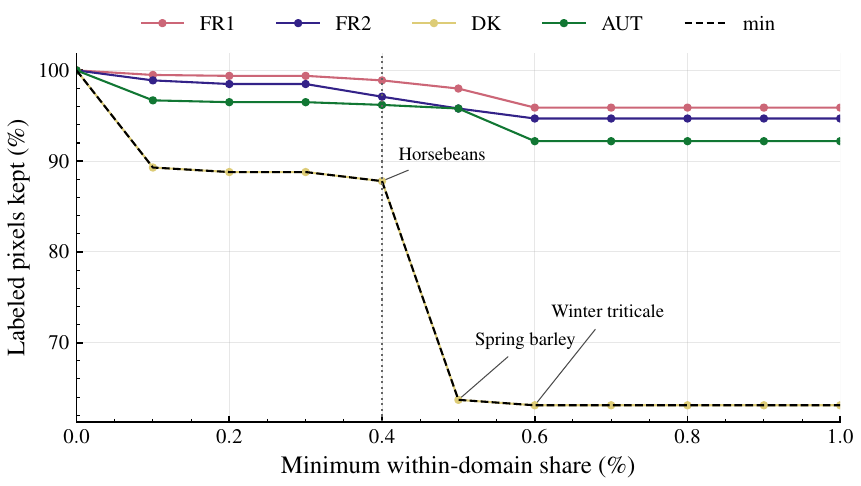}
    \caption{Share of labeled pixels kept in each \cropdataname domain as a function of the class-retention threshold. A class is kept if it covers at least this share of the labeled pixels in each of the four domains. The chosen threshold is the dotted line at 0.4\%.}
    \label{fig:tresh_crops}
\end{figure}

The threshold is the largest value that retains Spring Barley, the dominant spring crop in Denmark (24.2\% of its labeled pixels), whose share is only 0.41\% in AUT (\cref{fig:tresh_crops}). At 0.5\%, Spring Barley is discarded and the share of Danish labels kept drops from 87.8\% to 63.7\%, while the three other domains stay above 95\%. With the chosen threshold, 98.9\%, 97.1\%, 87.8\%, and 96.2\% of the labeled pixels are kept in FR1, FR2, DK, and AUT, respectively; the Danish loss comes mainly from Winter Rye (7.1\% of DK, below 0.2\% in both French domains).

\subsection{Hyperparameters}
\label[appendix]{app:hyperparameters}

\subsubsection{Pre-training}
The learning rate increases linearly from $10^{-5}$ to $10^{-4}$ during the first 20 epochs (5\% of training), then decreases to $10^{-8}$ following a cosine schedule.
The scheduler is updated at each epoch.

\subsection{In-domain patterns}
\label[appendix]{app:indomain}
\subsubsection{Douglas in PolFI and PF}
\begin{figure}
    \includegraphics[width=0.9\linewidth]{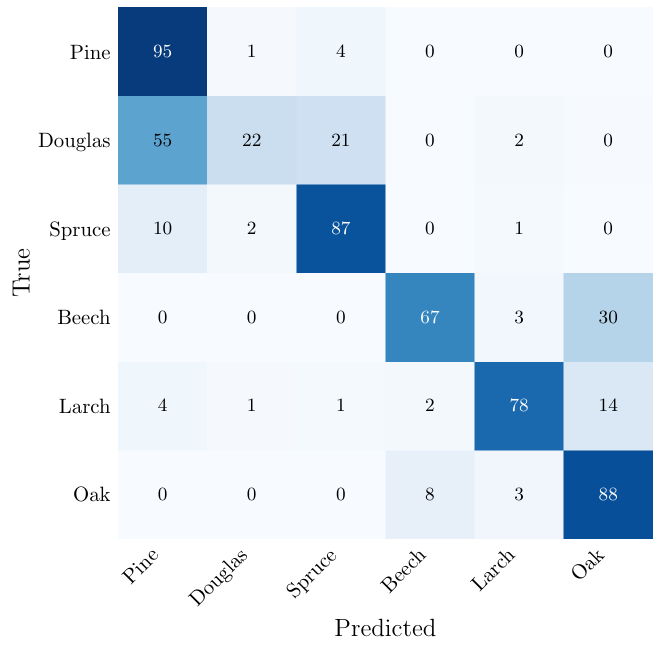} 
\caption{In-domain confusion matrices on PolFI (row-normalized, \%),
    averaged over ALISE, AnySat, TESSERA-S2, and \modelname.}
    \label{fig:cm_polfi}
\end{figure}

\begin{figure} 
    \includegraphics[width=0.9\linewidth]{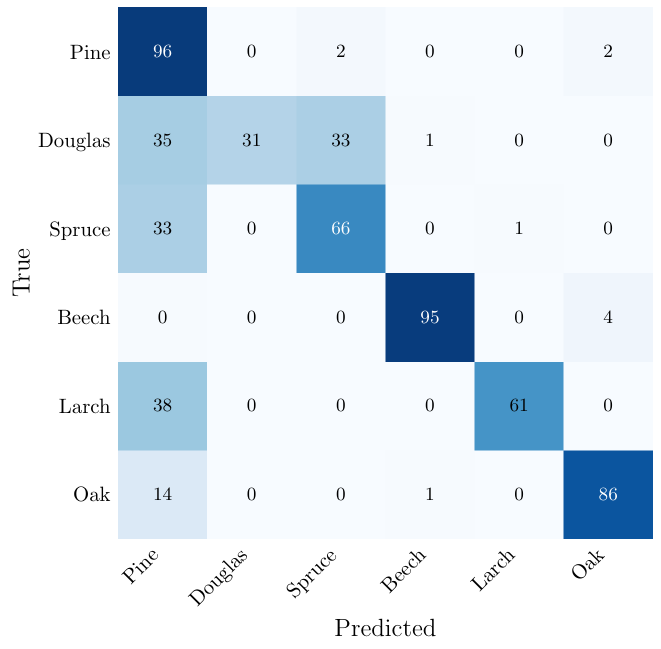} 
\caption{In-domain confusion matrices on PF (row-normalized, \%),
    averaged over ALISE, AnySat, TESSERA-S2, and \modelname.}
        \label{fig:cm_pf}
\end{figure}

\label[appendix]{app:polfi_pine_doug}
\Cref{fig:cm_polfi,fig:cm_pf} show the in-domain confusion matrices of PolFI and PF, averaged over ALISE, AnySat, TESSERA-S2, and \modelname. In both domains, the errors on Douglas stay within the evergreen conifers: in PolFI it is mostly predicted as Pine (55\%) and Spruce (21\%), in PF as Pine (35\%) and Spruce (33\%), and almost never as a broadleaf class.

\subsubsection{Douglas in PF across encoders}
\begin{figure} 
    \centering
    \includegraphics[width=0.9\linewidth]{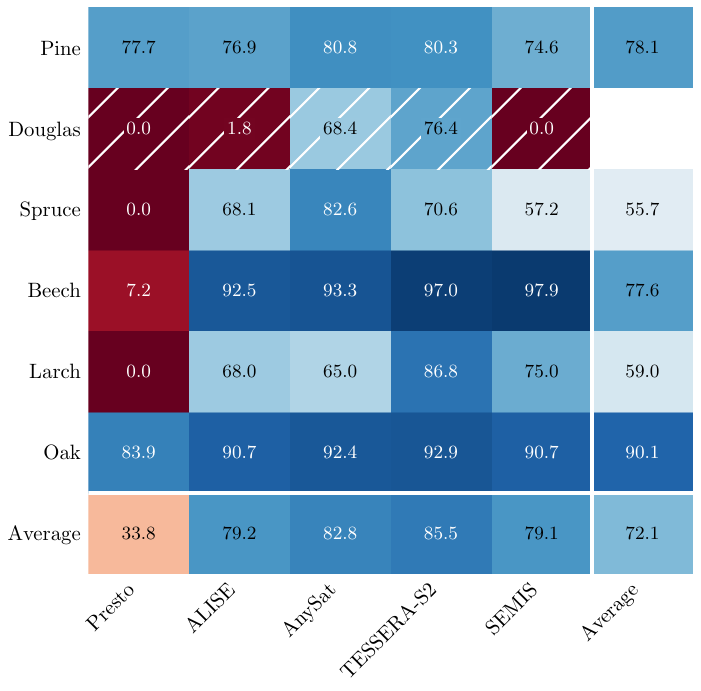}
    \caption{In-domain F1 per class and encoder on PF. The Average column excludes Douglas (hatched, difficult class), with a color scale centered on 50.}
    \label{fig:pf_ind}
\end{figure}

\label[appendix]{app:pf_ind}
\Cref{fig:pf_ind} reports the in-domain F1 of each encoder on PF. The Douglas class splits the encoders into two groups: ALISE and \modelname nearly fail (1.8 and 0 F1), whereas AnySat and TESSERA-S2 reach 68.4 and 76.4. Douglas is the scarcest class of PF (\cref{fig:support_sp}).

\subsubsection{Crop in-domain confusions}
\begin{figure} 
    \centering
    \includegraphics[width=0.9\linewidth]{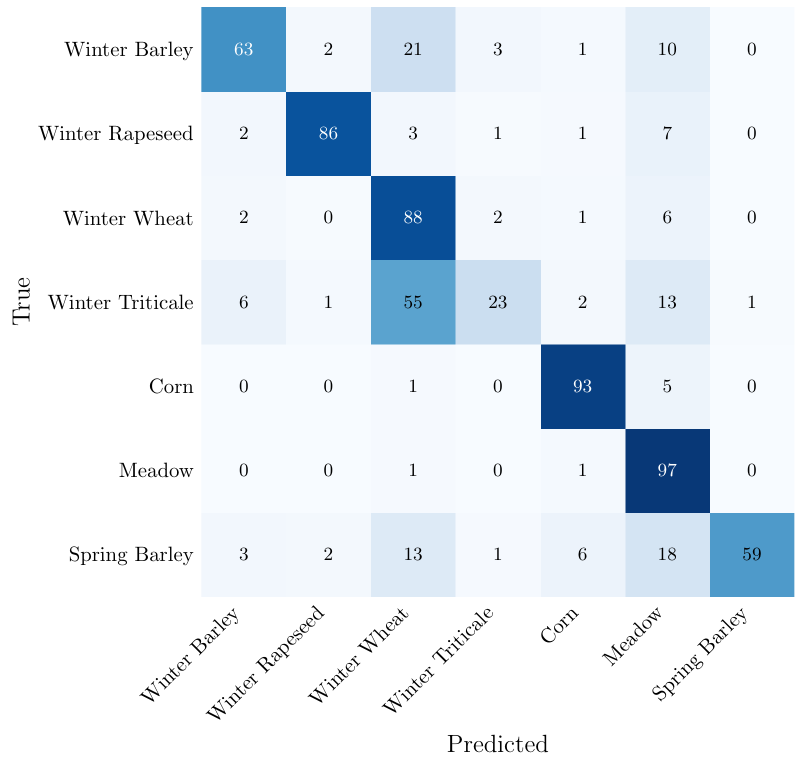}
    \caption{In-domain confusion matrices on \cropdataname (row-normalized, \%), FR1.}
    \label{fig:cm_crops_in_FR1}
\end{figure}

\begin{figure} 
    \centering
    \includegraphics[width=0.9\linewidth]{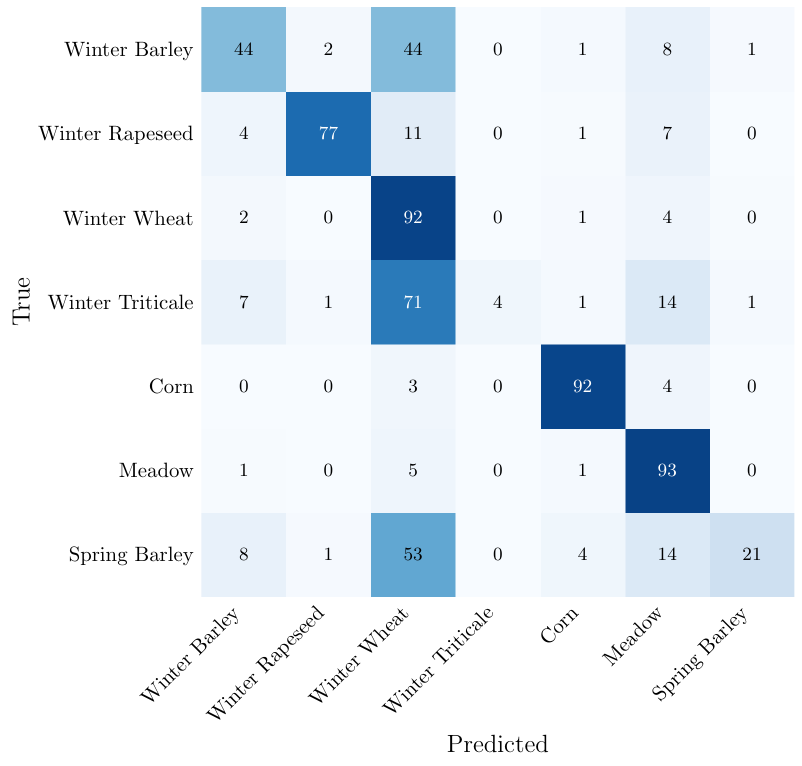}
    \caption{In-domain confusion matrices on \cropdataname (row-normalized, \%), FR2.}
    \label{fig:cm_crops_in_FR2}
\end{figure}

\begin{figure} 
    \centering
    \includegraphics[width=0.9\linewidth]{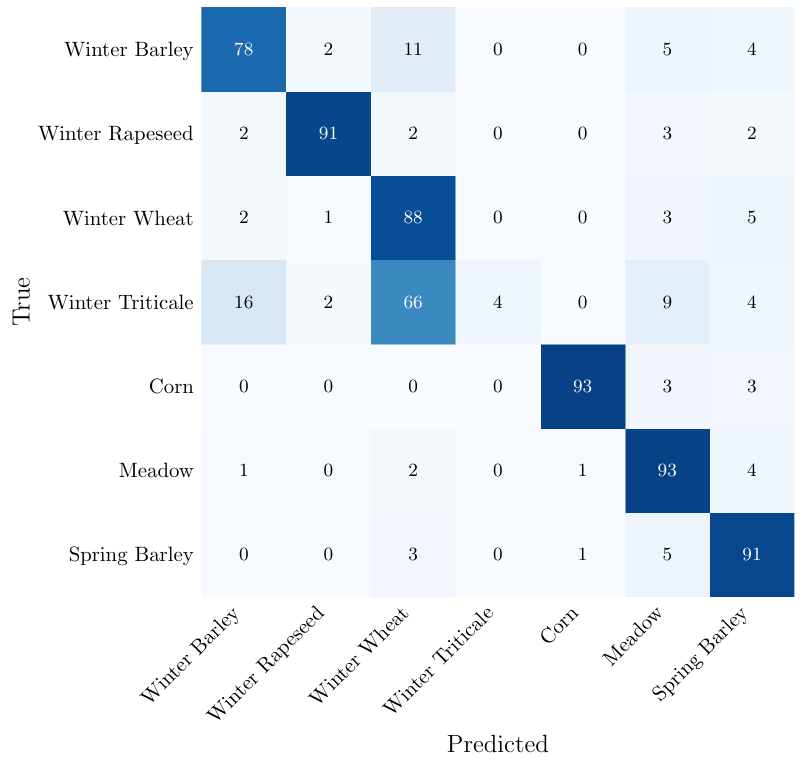}
    \caption{In-domain confusion matrices on \cropdataname (row-normalized, \%), DK.}
    \label{fig:cm_crops_in_DK}
\end{figure}

\begin{figure} 
    \centering
    \includegraphics[width=0.9\linewidth]{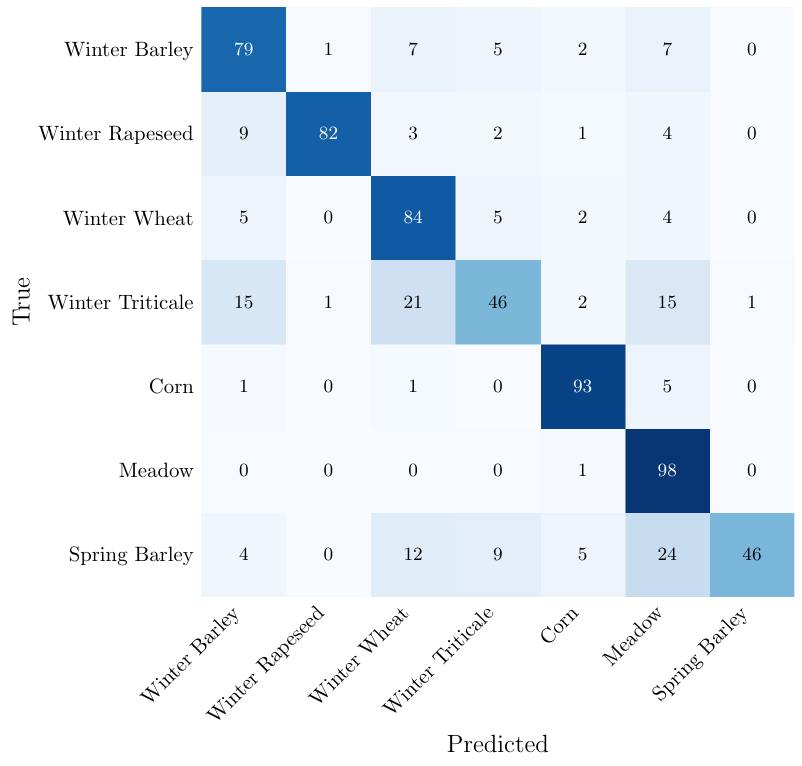}
    \caption{In-domain confusion matrices on \cropdataname (row-normalized, \%), AUT. Averaged over ALISE, AnySat, TESSERA-S2, and \modelname.}
    \label{fig:cm_crops_in_AUT}
\end{figure}

\label[appendix]{app:crop-conf}
\Cref{fig:cm_crops_in_FR1,fig:cm_crops_in_FR2,fig:cm_crops_in_DK,fig:cm_crops_in_AUT} show the in-domain confusion matrices of the four \cropdataname domains. In every domain, the main confusion of Winter Triticale is Winter Wheat (55\%, 71\%, 66\%, and 21\% of its pixels in FR1, FR2, DK, and AUT). Winter Barley is mostly confused with Winter Wheat in FR1 (21\%) and FR2 (44\%), much less in DK (11\%) and AUT (7\%). Spring Barley, well classified in DK (91\%), is mainly predicted as Winter Wheat in FR2 (53\%) and as Meadow in AUT (24\%).

\subsection{Auxiliary variables as pre-training targets}
\begin{figure} 
    \centering
    \includegraphics[width=\linewidth]{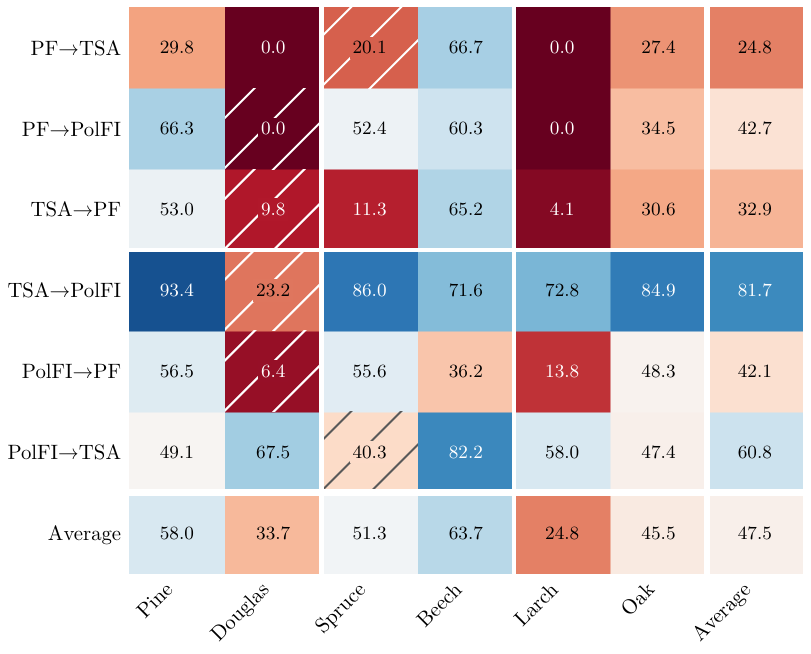}
    \caption{Cross-domain F1 of \outtopo per class and direction on \specdataname. Hatched cells are difficult classes of the target and are excluded from the Average row and column, with a color scale centered on 50.}
    \label{fig:topo_sp}
\end{figure}

\label[appendix]{app:aux-topo}
\Cref{fig:topo_sp} reports the cross-domain F1 of \outtopo{} per class and direction on \specdataname. On TSA$\to$PolFI, the classes that are not difficult in the target domain reach 81.7 F1 on average, compared with 64.7 for \modelname. The directions involving PF remain low (24.8 to 42.7), in line with the annotation mismatch discussed in \cref{subsec:bench_cross}.

\subsubsection{Recovered classes}
\label[appendix]{app:recovered}
\Cref{app:recovered_sp,app:recovered_tm} detail, for each class/direction pair identified as a cross-domain pattern in \cref{subsec:bench_cross}, the F1 of \modelname, of the best baseline model, and of \modelnameout. These pairs are, for \cropdataname, the 23 pairs whose F1 averaged over the encoders falls below 50 (\cref{fig:cross_f1_class_tm}), and for \specdataname, the 10 retained class/direction pairs of TSA$\leftrightarrow$PolFI. For each pair, we report the \modelnameout variant with the largest gain over \modelname, and the best baseline model is the best of ALISE, AnySat, and TESSERA-S2 on that pair. \Cref{tab:recovered} summarizes these tables per class.
 
On \specdataname (\cref{app:recovered_sp}), \modelnameout exceeds the best baseline model on all 10 pairs and improves over \modelname on 9 of them; the only exception, Pine on PolFI$\to$TSA, stays at the same level (49.1 versus 49.3). The selected auxiliary source is topography on 7 of the 10 pairs, consistent with the domain-level results of \cref{subsec:aux_res}.
 
On \cropdataname (\cref{app:recovered_tm}), \modelnameout improves over \modelname on 19 of the 23 pairs and exceeds the best baseline model on 14. The selected source follows the pair: soil or era5 on the seven pairs from DK, and topography or the combination on most pairs involving AUT. Three pairs remain low: Winter Barley on FR1$\to$DK (22.6 F1), and Winter Wheat on DK$\to$FR1 and DK$\to$FR2 (41.4 and 28.8 F1), where TESSERA-S2 remains well ahead (57.1 and 72.3 F1). Four pairs are degraded with respect to \modelname, three of them with AUT as source.
 
\input{sections/tables/recovered_sp}
\input{sections/tables/recovered_tm}

%% file: sections/tables/spec_mapping.tex
\providecommand{\greymidrule}{\arrayrulecolor{black!30}\midrule\arrayrulecolor{black}}
\providecommand{\na}{--}

\begin{table*}[]
\centering
\caption{\scriptsize{Harmonization of the three national nomenclatures into the six SPAN classes. Each cell gives the class label as published by the dataset authors and, in italics, the taxa it covers (TSA labels are taxon names). \na: class absent from the dataset.}}
\label{tab:span-harmonization}
\renewcommand{\cellalign}{lt}
\scriptsize    
\begin{tabular}{lllll}
\toprule
\textbf{Group} & \textbf{SPAN class} & \textbf{PolFI} \cite{grabska2024map} & \textbf{TSA} \cite{ahlswede2022treesatai} & \textbf{PF} \cite{gaydon2025pureforest} \\
\midrule
\multicolumn{5}{l}{\textbf{\textit{Retained in SPAN}}}\\
\midrule
\makecell{Evergreen\\ needleleaf} & \textbf{Pine} (\textit{Pinus})
  & \makecell{Pine (\textit{P.~sylvestris})\\ Dwarf mountain pine (\textit{P.~mugo})}
  & \makecell{\textit{P.~sylvestris}\\ \textit{P.~nigra}\\ \textit{P.~strobus}}
  & \makecell{Scotch pine (\textit{P.~sylvestris})\\ Black pine (\textit{P.~nigra})\\ Maritime pine (\textit{P.~pinaster})\\ Aleppo pine (\textit{P.~halepensis})} \\
\greymidrule
 & \textbf{Spruce} (\textit{Picea})
  & Spruce (\textit{P.~abies}) & \textit{P.~abies} & Spruce (\textit{P.~abies}) \\
\greymidrule
 & \makecell{\textbf{Douglas} (\textit{Pseudotsuga})}
  & \makecell{Douglas fir  (\textit{P.~menziesii})} & \textit{P.~menziesii} & Douglas (\textit{P.~menziesii}) \\
\greymidrule
\makecell{Deciduous needleleaf} & \textbf{Larch} (\textit{Larix})
  & Larch (\textit{L.~decidua})
  & \makecell{\textit{L.~decidua}\\ \textit{L.~kaempferi}}
  & Larch (\textit{L.~decidua}) \\
\midrule
\makecell{Deciduous broadleaf} & \textbf{Beech} (\textit{Fagus})
  & Beech (\textit{F.~sylvatica}) & \textit{F.~sylvatica} & Beech (\textit{F.~sylvatica}) \\
\greymidrule
 & \textbf{Oak} (\textit{Quercus})
  & Oak (\textit{Q.~robur}, \textit{Q.~petraea})
  & \makecell{\textit{Q.~robur}\\ \textit{Q.~petraea}\\ \textit{Q.~rubra}}
  & \makecell{Deciduous oak \\ (\textit{Q.~robur}, \textit{Q.~petraea}, \\ \textit{Q.~pubescens}, \textit{Q.~rubra})} \\
\midrule
\multicolumn{5}{l}{\textbf{\textit{Discarded but shared between PolFI and TSA}}}\\
\midrule
Deciduous broadleaf & \makecell{Alder, Birch,\\ Ash, Poplar}
  & \makecell{Alder, Birch,\\ Ash, Poplar}
  & \makecell{\textit{Alnus} sp., \textit{Betula} sp.,\\ \textit{F.~excelsior},\\ \textit{Populus} sp.}
  & \na \\
\midrule
\multicolumn{5}{l}{\textbf{\textit{Discarded}}}\\
\midrule
Evergreen needleleaf & Fir (\textit{Abies})$^{a}$
  & Fir (\textit{A.~alba}) & \textit{A.~alba} & \makecell{Fir (\textit{A.~alba}, \textit{A.~nordmanniana})} \\
\greymidrule
Deciduous broadleaf & Maple (\textit{Acer}) & Maple & \textit{A.~pseudoplatanus} & \na \\
 & Lime (\textit{Tilia})           & Lime         & \textit{Tilia} sp. & \na \\
 & Black locust (\textit{Robinia}) & Black locust & \na               & Black locust \\
 & Hornbeam (\textit{Carpinus})    & Hornbeam     & \na               & \na \\
 & Cherry (\textit{Prunus})        & \na          & \textit{Prunus} sp. & \na \\
 & Chestnut (\textit{Castanea})    & \na          & \na               & Chestnut \\
\greymidrule
Evergreen  broadleaf & Evergreen Oak (\textit{Q.~ilex}) & \na & \na & Evergreen oak \\
\greymidrule
Other & Cleared & \na & Cleared & \na \\
\midrule
& \# published classes & 16 & 20 & 13 \\
\bottomrule
\multicolumn{5}{p{0.85\textwidth}}{\footnotesize
$^{a}$~Present in the three datasets but not retained. In preliminary experiments with SEMIS, Fir obtained an in-domain F1 of 0 on PF, where 88\% of its pixels were predicted as Spruce, it would therefore have been a difficult class (\cref{subsec:benchin_dom}). Merging it with Douglas fir into a single class lowered the PF in-domain F1 of that class to 35, so Douglas fir was kept alone even if it emerged as an outlier in the domain in the following experiments.}
\end{tabular}
\end{table*}

%% file: sections/tables/recovered_sp.tex
\begin{table}[H]
\caption{Per-pair effect of \modelnameout on the TSA$\leftrightarrow$PolFI pairs of \specdataname, excluding difficult classes of the target. Only the best \modelnameout variant of each pair is reported. Green: \modelnameout outperforms the best baseline model.}
\label{app:recovered_sp}
\setlength{\tabcolsep}{4pt}
\begin{tabular}{l|lc|rrr}
\toprule
& & Auxiliary&\multicolumn{3}{c}{F1} \\
Class & Direction &  Source & \modelname & Best base. & \modelnameout  \\
\midrule
\multirow[t]{2}{*}{Beech} & PolFI$\to$TSA & soil & 71.3 & 75.1 & \cellcolor{secondgreen}\textbf{83.4} \\
 & TSA$\to$PolFI & combo & 66.8 & 70.2 & \cellcolor{secondgreen}\textbf{75.5} \\
\grouprule{1-6}
Douglas & PolFI$\to$TSA & topo & 52.1 & 57.0 & \cellcolor{secondgreen}\textbf{67.5} \\
\grouprule{1-6}
\multirow[t]{2}{*}{Larch} & PolFI$\to$TSA & topo & 51.9 & 50.6 & \cellcolor{secondgreen}\textbf{58.0} \\
 & TSA$\to$PolFI & topo & 44.6 & 67.8 & \cellcolor{secondgreen}\textbf{72.8} \\
\grouprule{1-6}
\multirow[t]{2}{*}{Oak} & PolFI$\to$TSA & era5 & 43.4 & 43.1 & \cellcolor{secondgreen}\textbf{50.1} \\
 & TSA$\to$PolFI & topo & 76.0 & 71.8 & \cellcolor{secondgreen}\textbf{84.9} \\
\grouprule{1-6}
\multirow[t]{2}{*}{Pine} & PolFI$\to$TSA & topo & 49.3 & 48.8 & \cellcolor{secondgreen}\textbf{49.1} \\
 & TSA$\to$PolFI & topo & 80.7 & 87.7 & \cellcolor{secondgreen}\textbf{93.4} \\
\grouprule{1-6}
Spruce & TSA$\to$PolFI & topo & 55.5 & 71.6 & \cellcolor{secondgreen}\textbf{86.0} \\
\bottomrule
\end{tabular}
\end{table}

%% file: sections/tables/recovered_tm.tex
\begin{table}[H]
\caption{Per-pair effect of \modelnameout on the \cropdataname class/direction pairs identified as pattern (ii). Only the best \modelnameout variant of each pair is reported. Green: \modelnameout outperforms the best baseline model.}
\label{app:recovered_tm}
\setlength{\tabcolsep}{3pt}
\begin{tabular}{l|lc|rrr}
\toprule
& & Auxiliary&\multicolumn{3}{c}{F1} \\
Class & Direction &  Source & \modelname & Best base. & \modelnameout  \\
\midrule
\multirow[t]{11}{*}{Winter Barley} & FR1$\to$FR2 & era5 & 48.3 & 38.3 & \cellcolor{secondgreen}\textbf{50.4} \\
 & FR1$\to$DK & era5 & 0.3 & 46.3 & 22.6 \\
 & FR1$\to$AUT & era5 & 25.1 & 56.9 & 44.4 \\
 & FR2$\to$FR1 & topo & 59.0 & 46.2 & \cellcolor{secondgreen}\textbf{65.0} \\
 & FR2$\to$DK & era5 & 0.0 & 12.7 & \cellcolor{secondgreen}\textbf{13.5} \\
 & FR2$\to$AUT & era5 & 11.7 & 34.0 & 24.1 \\
 & DK$\to$FR1 & soil & 25.2 & 33.9 & 29.0 \\
 & DK$\to$FR2 & soil & 18.1 & 28.2 & 22.2 \\
 & AUT$\to$FR1 & combo & 39.5 & 28.9 & \cellcolor{secondgreen}\textbf{37.1} \\
 & AUT$\to$FR2 & topo & 28.1 & 28.6 & \cellcolor{secondgreen}\textbf{29.0} \\
 & AUT$\to$DK & era5 & 41.2 & 52.8 & \cellcolor{secondgreen}\textbf{70.1} \\
\grouprule{1-6}
\multirow[t]{3}{*}{Winter Wheat} & DK$\to$FR1 & soil & 5.2 & 57.1 & 41.4 \\
 & DK$\to$FR2 & soil & 3.9 & 72.3 & 28.8 \\
 & AUT$\to$FR1 & combo & 49.7 & 48.9 & \cellcolor{secondgreen}\textbf{65.1} \\
\grouprule{1-6}
\multirow[t]{4}{*}{Winter Rapeseed} & FR1$\to$AUT & combo & 36.2 & 62.6 & \cellcolor{secondgreen}\textbf{73.8} \\
 & FR2$\to$AUT & topo & 51.1 & 33.1 & \cellcolor{secondgreen}\textbf{67.2} \\
 & DK$\to$FR2 & era5 & 49.4 & 47.4 & 44.6 \\
 & AUT$\to$FR2 & era5 & 36.5 & 44.5 & 35.6 \\
\grouprule{1-6}
\multirow[t]{4}{*}{Spring Barley} & FR2$\to$FR1 & soil & 44.9 & 25.4 & \cellcolor{secondgreen}\textbf{55.3} \\
 & FR2$\to$DK & combo & 3.9 & 26.0 & \cellcolor{secondgreen}\textbf{58.6} \\
 & DK$\to$FR1 & soil & 20.7 & 14.8 & \cellcolor{secondgreen}\textbf{40.4} \\
 & AUT$\to$FR1 & topo & 60.2 & 36.1 & \cellcolor{secondgreen}\textbf{45.6} \\
\grouprule{1-6}
Corn & DK$\to$FR1 & era5 & 21.6 & 63.9 & \cellcolor{secondgreen}\textbf{79.8} \\
\bottomrule
\end{tabular}
\end{table}